\documentclass[final,5p,times,twocolumn]{elsarticle}

\usepackage[utf8]{inputenc}
\usepackage[T1]{fontenc}

\usepackage{amssymb}

\usepackage{amsfonts}
\usepackage{amsmath}
\usepackage{algorithmic}
\usepackage{xcolor}
\usepackage{graphicx}
\usepackage{tabularx}
\usepackage{pgfplots}
\pgfplotsset{compat=1.18}	
\usetikzlibrary{pgfplots.statistics} 
\usetikzlibrary{matrix}
\usepgfplotslibrary{groupplots}
\usepackage{lipsum}
\usepackage{bbm}

\usepackage{tikz}
\usetikzlibrary{positioning, calc, chains}
\usetikzlibrary{decorations, decorations.pathreplacing, decorations.pathmorphing, calligraphy}
\usetikzlibrary{arrows.meta}
\usetikzlibrary{
 shapes,
 shapes.geometric,
 shapes.symbols,
 shapes.arrows,
 shapes.multipart,
 shapes.callouts,
 shapes.misc}

\usepackage{hyperref}
\usepackage{array}
\usepackage{multirow}
\usepackage{color, colortbl}
\usepackage{comment}
\usepackage{booktabs}

\hypersetup{
 colorlinks=false,
 linkbordercolor=white,
 urlbordercolor=white,
 pdfborder={0 0 0}
}

\newcommand{\bl}{\color{blue}}
\newcommand{\bb}{\color{black}}

\journal{International Journal of Information Management}

\begin{document}

\begin{frontmatter}
 
\title{Resilient Vision-Tabular Multimodal Learning under Modality Missingness}

\author[UCBM]{Camillo Maria Caruso} 
\ead{camillomaria.caruso@unicampus.it}

\author[UCBM]{Valerio Guarrasi\fnref{eq}} 
\ead{valerio.guarrasi@unicampus.it}

\author[UCBM,UMU]{Paolo Soda\fnref{eq}\corref{cor1}} 
\ead{p.soda@unicampus.it, paolo.soda@umu.se}
\cortext[cor1]{Corresponding author: paolo.soda@umu.se}
\fntext[eq]{These authors contributed equally to this work.}

\affiliation[UCBM]{organization={Research Unit of Artificial Intelligence and Computer Systems, Department of Engineering, Università Campus Bio-Medico di Roma},
 city={Roma},
 state={Italy},
 country={Europe}}

\affiliation[UMU]{organization={Department of Diagnostics and Intervention, Radiation Physics, Biomedical Engineering, Umeå University},
 city={Umeå},
 state={Sweden},
 country={Europe}}

\begin{abstract}

Multimodal deep learning has shown strong potential in medical applications by integrating heterogeneous data sources such as medical images and structured clinical variables. 
However, most existing approaches implicitly assume complete modality availability, an assumption that rarely holds in real-world clinical settings where entire modalities and individual features are frequently missing. 
In this work, we propose a multimodal transformer framework for joint vision-tabular learning explicitly designed to operate under pervasive modality missingness, without relying on imputation or heuristic model switching.
The architecture integrates three components: a vision, a tabular, and a multimodal fusion encoder.
Unimodal representations are weighted through learnable modality tokens and fused via intermediate fusion with masked self-attention, which excludes missing tokens and modalities from information aggregation and gradient propagation. 
To further enhance resilience, we introduce a modality-dropout regularization strategy that stochastically removes available modalities during training, encouraging the model to exploit complementary information under partial data availability.
We evaluate our approach on the MIMIC-CXR dataset paired with structured clinical data from MIMIC-IV for multilabel classification of 14 diagnostic findings with incomplete annotations. 
Two parallel systematic stress-test protocols progressively increase training and inference missingness in each modality separately, spanning fully multimodal to fully unimodal scenarios. 
Across all missingness regimes, the proposed method consistently outperforms representative baselines, showing smoother performance degradation and improved robustness. 
Ablation studies further demonstrate that attention-level masking and intermediate fusion with joint fine-tuning are key to resilient multimodal inference.
\end{abstract}


\end{frontmatter}

\section{Introduction}\label{sec:introduction}
In recent years, multimodal deep learning has become a cornerstone of artificial intelligence (AI) in healthcare, enabling the integration of heterogeneous data such as medical images, clinical variables, and laboratory results to provide more accurate and interpretable diagnostic and prognostic tools~\cite{bib:review_intermediate}. 
By combining complementary sources of information, multimodal models can capture complex relationships that remain hidden in unimodal analyses, ultimately improving clinical decision-making~\cite{bib:review_multimodal, bib:review_multimodal2}.

Most multimodal AI systems implicitly assume that every patient record contains data from all the required modalities but, in practice, this assumption rarely holds~\cite{bib:review_multimodal}. 
Missing information, whether individual features or entire modalities, is a pervasive issue caused by several factors, such as data fragmentation across institutions, incompatible storage platforms, retrospective collected data, or privacy limitations. 

However, missing data can seriously compromise model reliability and generalization, especially when this missingness affects modalities that are usually more informative for making a decision, such as medical imaging~\cite{bib:review_multimodal_challenges}. 
Therefore, multimodal deep learning models must be explicitly designed to be resilient, meaning that they can be trained and perform reliably even under partial data availability, without relying on weak assumptions of complete inputs.
Hereinafter, we use the term missing data to indicate scenarios in which, for a given patient, an entire modality is unavailable. 
Although the proposed framework is capable of handling both forms of incompleteness, either individual features within a modality or entire modalities, the primary focus of this work is on resilience to missing modalities, which represents a particularly critical and understudied challenge in real-world clinical settings.

Classical approaches to address missing data rely on modality completion techniques, which attempt to reconstruct absent information from the available one~\cite{bib:review_missing}. 
While simple to implement, these methods may introduce bias by inferring values that do not faithfully reflect the patient’s actual underlying measurements, degrading downstream performance and driving the model toward clinically misleading predictions. 
To avoid these issues, recent research has focused on the following four main strategies that make models inherently resilient to missing inputs~\cite{bib:review_missing}:
(i) \textit{Modality Augmentation} methods, which replace missing representations, usually with constant or random values~\cite{bib:modality_augmentation};
(ii) \textit{Feature Space Engineering} techniques, which combine unimodal representations through simple operations such as averaging or pooling~\cite{bib:feature_space_engineering};
(iii) \textit{Architecture Engineering} approaches, which explicitly model modality-wise interactions and missingness (e.g., via attention mechanisms)~\cite{bib:architecture_engineering}; and
(iv) \textit{Model Selection} methods, which dynamically switch between unimodal and multimodal predictors based on the modalities available at inference time~\cite{bib:model_selection}.
While the interested readers can refer to~\cite{bib:review_missing} for further details, on the one hand, we notice that the approaches belonging to categories (i), (ii) and (iv) rely on simplified fusion schemes or modality-wise heuristics, treating the modalities as loosely coupled sources, thus preventing the model from learning the fine-grained cross-modal dependencies that are essential for robust clinical reasoning. 
On the other hand, the methods from group (iii) exploit cross-modality supervision during pre-training to enhance downstream unimodal performance.
Representative approaches are TIP~\cite{bib:TIP} and CHARM~\cite{bib:CHARMS}, which jointly learn from image-tabular pairs to infuse the unimodal vision encoder with complementary cross-modal information, thereby improving its representation power and performance.
While effective in enhancing unimodal representations, they do not address multimodal training and inference under arbitrary modality availability, as they remain dependent on the presence of the target modality.
Because real-world clinical applications often lack complete imaging or tabular information, their applicability becomes constrained outside of controlled benchmark settings.
This motivates the need for architectures that can operate natively on any subset of available modalities and sustain reliable performance even under extreme and unpredictable conditions with missing modalities.

To address the issue of \textit{Multimodal Learning with Missing Modalities} described so far, here we propose a multimodal vision-tabular transformer framework designed for flexible data fusion.
Rather than attempting to reconstruct absent inputs, the model conditions its training and inference exclusively on the information that is actually available, allowing it to adapt seamlessly to any subset of modalities.
The architecture performs intermediate fusion through a unified transformer encoder trained end-to-end, enabling the model to learn joint representations that integrate imaging and tabular information.
Moreover, to prevent the model from over-relying on the most informative modality and to promote balanced learning across sources, we introduce a modality-dropout regularization strategy that randomly removes one of the available modalities during training. 
This mechanism forces the model to exploit complementary information and discourages degenerate solutions dominated by a single modality.
Additionally, our method not only enables resilience when an entire modality is absent, but also naturally accommodates fine-grained missingness within the tabular domain, a widespread circumstance in real settings~\cite{bib:MIMIC_review}. 

To rigorously evaluate the proposed framework, we formulate two complementary research questions (RQs) that capture distinct challenges encountered in clinical multimodal learning.
The first question examines how effectively the model can be trained when multimodal data are only partially available. 
This situation commonly arises in retrospective or multi-center datasets, where different data sources are collected at different times or only for subsets of patients~\cite{bib:MIMIC_review}. 
As a result, learning must proceed from incomplete and weakly paired supervision, requiring the model to extract meaningful representations even when cross-modal alignment is limited.
The second question focuses on the model’s behavior at inference time under varying levels of data availability. 
In real deployment scenarios, one or more modalities may be missing due to acquisition delays, limited resources, or fragmented records. Under these conditions, the model must adapt its predictions to the subset of observed inputs while maintaining stable and reliable performance.

To address these aspects, we evaluate our approach on the MIMIC-CXR dataset~\cite{bib:MIMIC-CXR} paired with clinical records from MIMIC-IV~\cite{bib:MIMIC-IV}, and we design two controlled evaluation protocols that progressively reduce the availability of each modality, either during training or only at test time. 
This setup allows us to disentangle the impact of incomplete supervision from that of partial observations at inference, and to assess robustness across the entire spectrum of data availability, from fully multimodal inputs to strictly unimodal scenarios.

The main contributions of this work are therefore summarized as follows:
\begin{itemize}
\item Multimodal architecture for incomplete data: We propose a multimodal vision-tabular transformer that, through a masked attention mechanism, is resilient during training and inference to any missing modality. 

\item Robust evaluation: We evaluate our approach on 62071 patient records from the MIMIC-CXR dataset, comparing also with three state-of-the-art approaches representing the main families of missing-modality strategies, as well as early and late fusion baselines.

\item Interpretability under missingness: We provide an interpretability analysis of modality importance under increasing missingness, revealing how complementary modalities contribute to decision-making.
\end{itemize}

To support reproducibility and facilitate comparison with future methods, the implementation of the proposed framework and the experimental protocols is publicly available at: \url{https://github.com/arco-group/Resilient-VT}.

The remainder of this paper is organized as follows. 
Section~\ref{sec:methods} presents the proposed multimodal framework, detailing the vision, tabular, and multimodal encoders, together with the masking mechanisms, modality-dropout regularization, and the training strategy. 
Section~\ref{sec:material} describes the dataset construction and preprocessing pipeline, including the alignment of imaging and clinical data. 
Section~\ref{sec:competitors} introduces the experimental design, including the considered baselines and the ablation analyses used to assess the contribution of each component. 
Section~\ref{sec:experiments} details the experimental configuration and evaluation protocols adopted to study robustness under varying data availability conditions. 
Section~\ref{sec:results} presents and discusses the results, structured around the two research questions, analyzing the model’s behavior under missingness in training and in test. 
Finally, Section~\ref{sec:conclusion} concludes the paper and outlines directions for future work.


\section{Methods}\label{sec:methods}

\begin{figure*}[!t]
\resizebox{\textwidth}{!}{
\begin{tikzpicture}
\node[label={[font=\small, align=center]below: (A)}] (A) at (0,0) {\includegraphics[width=.5\textwidth]{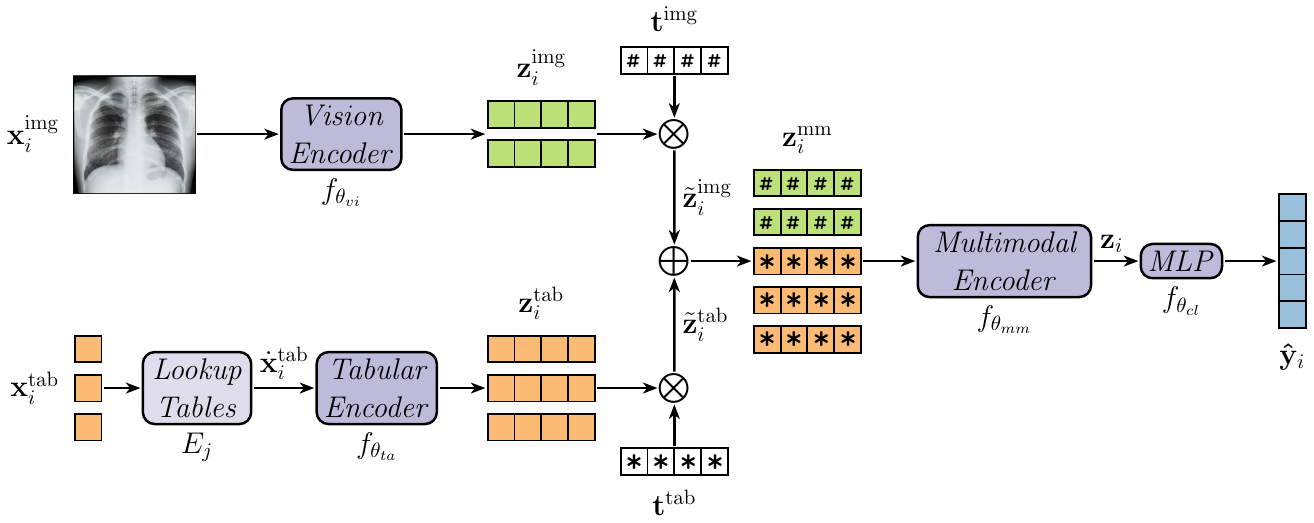}};
\node[label={[font=\small, align=center]below: (B)}] (B) [right=.1cm of A] {\includegraphics[width=.5\textwidth]{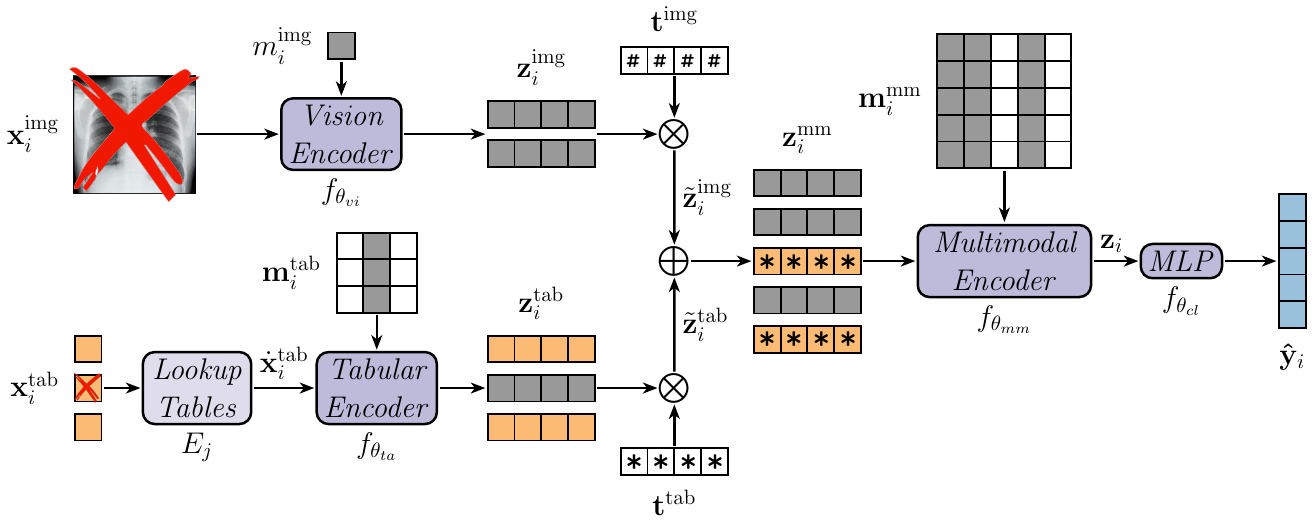}};
\draw[-] ($(A.south)!0.5!(B.south)$) -- ($(A.north)!0.5!(B.north)$);
\end{tikzpicture}}
\caption{Overview of the proposed multimodal architecture. (A) When both image and tabular data are available, the vision and tabular encoders produce modality-specific embeddings that are fused by the multimodal transformer. (B) Under missing-modality or missing-feature conditions, visual and tabular embeddings are masked, so the multimodal encoder integrates only the available information through masked attention.}\label{fig:model}
\end{figure*}

In the context of multimodal learning under heterogeneous and irregular data availability, where entire modalities, individual features and target labels may be missing at both training and inference time, let
\begin{equation}
\mathcal{D} = \{(\mathbf{x}_i^{\text{img}}, \mathbf{x}_i^{\text{tab}}, \mathbf{y}_i)\}_{i=1}^N
\end{equation}
be a multimodal dataset composed of $N$ patient records, where $\mathbf{x}_i^{\text{img}} \in \mathbb{R}^{H \times W}$ denotes the image of the $i$-th patient, $\mathbf{x}_i^{\text{tab}} \in \mathbb{R}^F$ the corresponding clinical variables, and $\mathbf{y}_i \in \{0,1\}^{C}$ the multilabel target vector across $C$ classes. 
To formulate multimodal learning as a function that is exclusively based on observed information, we introduce the following three binary masks that model when some instance is partially missing: 
\begin{itemize}
 \item $m_i^{\text{img}} \in \{0,1\}$, indicating whether the image modality is available (1) or missing (0);
 \item $\mathbf{m}_i^\text{tab} \in \{0,1\}^{F}$, denoting which of the $F$ tabular features are observed;
 \item $\mathbf{m}_i^\text{y} \in \{0,1\}^{C}$, specifying which of the $C$ multilabel classes are available.
\end{itemize}
Our model consists of three main components (\figurename~\ref{fig:model}): a vision encoder, a tabular encoder, and a multimodal fusion encoder. 
It learns a mapping
\begin{equation}
f_{\Theta} : (\mathbf{x}_i^{\text{img}}, \textbf{x}_i^{\text{tab}}, m_i^{\text{img}}, \mathbf{m}_i^{\text{tab}}) \longrightarrow \mathbf{\hat{y}}_i ,
\end{equation}
parameterized by $\Theta = \{\theta_{vi}, \theta_{ta}, \theta_{mm}, \theta_{cl}\}$, where the subscripts denote the parameters of the vision encoder $vi$, tabular encoder $ta$, multimodal encoder $mm$, and classification head $cl$, respectively; $\mathbf{\hat{y}}_i \in \mathbb{R}^C$ is the vector of estimated posterior class probabilities for the $i$-th patient. 
Since label annotations may also be partially missing, learning is performed exclusively on the subset of observed targets. 
The training objective is therefore defined through a masked multilabel binary cross-entropy loss that excludes unobserved labels from supervision. 
The complete formulation of the loss function and the training strategy is detailed in section~\ref{sec:loss}.

When both modalities are available ($\mathbf{x}_i^\text{img}$ and $\mathbf{x}_i^\text{tab}$ in \figurename~\ref{fig:model}.A), the vision and tabular encoders produce modality-specific latent representations ($\mathbf{z}_i^\text{img}$ and $\mathbf{z}_i^\text{tab}$), which are weighted using learnable modality tokens, $t^\text{img}$ and $t^\text{tab}$, and concatenated into a shared representation $\mathbf{z}_i^\text{mm}$ processed by the multimodal encoder. 
This enables intermediate fusion between the modalities, enabling joint optimization of unimodal and cross-modal representations.

Under partial data availability (\figurename~\ref{fig:model}.B), $m_i^{\text{img}} $ and $\mathbf{m}_i^{\text{tab}}$ are exploited by the model $f_{\Theta}$ to ensure that missing modalities and features do not contribute to representation learning, attention aggregation, or gradient propagation. 
Indeed, if the image modality is absent, the corresponding visual representation is multiplicatively masked by $m_i^\text{img}$, which prevents it from contributing to the fused representation and from influencing parameter updates in the vision branch, as it will be detailed in section~\ref{sec:vision_encoder}. 
Similarly, missing tabular features are handled through a masked self-attention mechanism using the mask $\mathbf{m}_i^\text{tab}$ (section~\ref{sec:tabular_encoder}).
Then, in the multimodal encoder (section~\ref{sec:multimodal_encoder}), both missing features or even entire missing modalities are masked using a similar self-attention mechanism that, based on the composite mask $\mathbf{m}_i^\text{mm}$, can suppress attention scores involving unobserved tokens, preventing missing features or modalities from either sending or receiving information. 
As a result, the multimodal encoder aggregates only the available evidence, dynamically adapting its reasoning process to the subset of available modalities for each patient.
To further promote robustness, modality-dropout regularization is applied during training (section~\ref{sec:modality_dropout}): 
by stochastically removing available modalities, the model is encouraged to learn complementary and non-redundant representations, enhancing its ability to generalize under severe and previously unseen missingness patterns.

\subsection{Vision Encoder}\label{sec:vision_encoder}
The vision encoder $f_{\theta_{vi}}$ extracts high-level visual representations from raw images $\mathbf{x}_i^{\text{img}} \in \mathbb{R}^{H \times W}$ using a residual convolutional backbone (section~\ref{sec:experimental_configuration}) producing a $d_v$-dimensional latent representation $\mathbf{z}_i^{\text{img}} $:
\begin{equation} \label{eq:VisionEncoder}
\mathbf{z}_i^{\text{img}} = f_{\theta_{vi}} (\mathbf{x}_i^{\text{img}};m_i^{\text{img}}) \in \mathbb{R}^{d_v}
\end{equation}

For compatibility with tabular tokens, this vector is reshaped into a $d$-dimensional latent representation, yielding $\mathbf{z}_i^{\text{img}} \in \mathbb{R}^{\frac{d_v}{d} \times d}$.
To ensure that the downstream multimodal module is aware of the modalities to which the input tokens belong, we introduce a learnable \textit{modality token} $\mathbf{t}^{\text{img}} \in \mathbb{R}^{d}$ that, together with $\mathbf{z}_i^{\text{img}}$, produces the conditioned latent representation $\mathbf{\tilde{z}}_i^{\text{img}}$ as follows:
\begin{equation}
\mathbf{\tilde{z}}_i^{\text{img}} = \mathbf{z}_i^{\text{img}} \odot \mathbf{t}^{\text{img}},
\end{equation}
where the $\odot$ denotes element-wise product.
When the image modality is missing, either during training or inference, this conditioned latent representation is masked out by the mask $m_i^{\text{img}}=0$, thus excluding it from gradient updates:
\begin{equation}
m_i^{\text{img}} \mathbf{\tilde{z}}_i^{\text{img}} = \mathbf{0} \quad \to \quad
\nabla_{\theta_{vi}}\mathbf{\tilde{z}}_i^{\text{img}} = \mathbf{0}.
\end{equation}
This mechanism ensures that missing images, since they do not need to be generated or replaced, neither contribute to the multimodal representation nor affect the learning dynamics of the visual branch.


\subsection{Tabular Encoder}\label{sec:tabular_encoder}
The tabular encoder $f_{\theta_{ta}}$ extracts feature-wise contextual representations from numerical and categorical clinical variables and transforms them into a sequence of latent representations suitable for attention-based processing:
\begin{equation} \label{eq:TabularEncoder}
\mathbf{z}_i^{\text{tab }} = f_{\theta_{ta}} (\mathbf{x}_i^{\text{tab}};\mathbf{m}_i^{\text{tab}}) \in \mathbb{R}^{F \times d}.
\end{equation}

To this aim, let first introduce the mapping we  apply to the feature vector $\mathbf{x}_i^{\text{tab}}$ so defined: 
\begin{equation}
\mathbf{x}_i^{\text{tab}} = [x_{i,1}^{\text{tab}}, \dots, x_{i,j}^{\text{tab}}, \dots, x_{i,F}^{\text{tab}}] \in \mathbb{R}^F,
\end{equation}
This operation maps each of its elements, considered as a single token, to the latent space through a feature-specific embedding function, producing the tabular embedding matrix $\dot{\mathbf{x}}_i^{\text{tab}}$:
\begin{equation}
\dot{\mathbf{x}}_i^{\text{tab}} = [\dot{\mathbf{x}}_{i,1}^{\text{tab}}, \dots, \dot{\mathbf{x}}_{i,j}^{\text{tab}}, \dots, \dot{\mathbf{x}}_{i,F}^{\text{tab}}] \in \mathbb{R}^{F \times d}.
\end{equation}
It is worth noting that we compute each row of this matrix, i.e., $\dot{\mathbf{x}}_{i,j}^{\text{tab}}$, using two different strategies according to the feature type:
\begin{equation}
\dot{\mathbf{x}}_{i,j}^{\text{tab}} = \left\{ 
\begin{aligned}
&E_{j_\text{cat}} (x_{i,j}^{\text{tab}}), \,\, \text{if } \, x_{i,j}^{\text{tab}} \,\,\, \text{is categorical} \\
x_{i,j}^{\text{tab}} &E_{j_\text{num}} (x_{i,j}^{\text{tab}}), \,\, \text{if } \, x_{i,j}^{\text{tab}} \,\,\, \text{is numerical}
\end{aligned}
\right.
\end{equation}
where both $E_{j_\text{cat}}$ and $E_{j_\text{num}}$ are look-up tables for categorical and numerical features. 
In other words, categorical features are mapped to indices and then embedded into a latent vector via $E_{j_{\text{cat}}}$, which assigns an embedding vector to each possible category.
Note that $E_{j_{\text{cat}}}$ contains an embedding vector for each possible entry in the specific categorical feature, and it is also able to encode missing features through a dedicated padding index related to a not-trainable embedding vector.
Numerical features, instead, are first normalized and their presence or missingness is encoded through $E_{j_\text{num}}$, which consists of two entries only (absence/presence). 
Then, when the feature is present, its value is multiplied by the embedding vector to maintain the feature's scale.

Turning back our attention to equation~\ref{eq:TabularEncoder}, $f_{\theta_{ta}}$ is a vanilla transformer-based encoder that models the  dependencies between the previous embeddings $\dot{\mathbf{x}}_i^{\text{tab}}$ through self-attention and produces the tabular latent representation $\mathbf{z}_i^{\text{tab}} \in \mathbb{R}^{F \times d}$.
To correctly cope with fine-grained missingness, the attention computation is guided by the binary mask $\mathbf{m}_i^\text{tab}$, ensuring that absent features do not influence the training process.
To enforce this behavior at the level of all query-key interactions, the mask vector is expanded into a matrix $\mathbf{m}_i^\text{tab} = \{0, 1\}^{F \times F}$ where each row replicates the feature-availability pattern.
We then introduce the following modified masked self-attention mechanism:
\begin{equation}\scalebox{.9}{$
\begin{aligned}
&\text{Attn}(\mathbf{q}_i^\text{tab}, \mathbf{k}_i^\text{tab}, \mathbf{v}_i^\text{tab}, \mathbf{m}_i^{\text{tab}}) = \\ &\text{ReLU}\left(\text{Softmax}\!\left(\frac{\mathbf{q}_i^\text{tab} (\mathbf{k}_i^\text{tab})^\text{T}}{\sqrt{d}} + \log(\mathbf{m}_i^{\text{tab}})\right) + \log(\mathbf{m}_i^{\text{tab}})^\text{T}\right)\mathbf{v}_i^\text{tab}.
\end{aligned}
$}
\end{equation}
where missing features are explicitly prevented from both sending and receiving attention.
Indeed, we modify the standard masked self-attention by adding the transpose of the masking matrix to the attention logits, together with a ReLU activation that suppresses any residual contributions involving missing features.
Finally, a modality token $\mathbf{t}^{\text{tab}} \in \mathbb{R}^d$ is multiplied element-wise by each latent representation in $\mathbf{z}_i^{\text{tab}}$, analogous to the one used in the vision branch, to scale the modality-level context. 
\begin{equation}
\mathbf{\tilde{z}}_i^{\text{tab}} = \mathbf{z}_i^{\text{tab}} \odot \mathbf{t}^{\text{tab}}.
\end{equation}


\subsection{Multimodal Encoder}\label{sec:multimodal_encoder}
The multimodal fusion encoder $f_{\theta_\text{mm}}$ integrates the latent representations produced by the vision and tabular encoders into a shared embedding space, enabling joint reasoning across modalities. 
Its design follows a transformer architecture equipped with a masked self-attention mechanism that selectively aggregates information from the available modalities, while we inject again the ability to ignore the missing ones.
Formally, 
\begin{equation} \label{eq:MultimodalEncoder}
\mathbf{z}_i = f_{\theta_\text{mm}} (\mathbf{z}_i^{\text{mm}};\mathbf{m}_i^{\text{mm}}) \in \mathbb{R}^{(\frac{d_v}{d} + F) \times d}
\end{equation}
where $\mathbf{z}_i^{\text{mm}}$ and $\mathbf{m}_i^{\text{mm}}$ are the multimodal latent representations and the composite mask, respectively.

The former, i.e., $\mathbf{z}_i^{\text{mm}}$, is the concatenation of the unimodal latent sequences along the token dimension:
\begin{equation}
\mathbf{z}_i^{\text{mm}} = [\mathbf{\tilde{z}}_i^{\text{img}}; \mathbf{\tilde{z}}_i^{\text{tab}}] \in \mathbb{R}^{(\frac{d_v}{d} + F) \times d}.
\end{equation}
This quantity is then processed by a stack of encoder layers from the vanilla transformer that model inter-modality dependencies through self-attention, finally returning the multimodal latent representation $\mathbf{z}_i \in \mathbb{R}^{(\frac{d_v}{d} + F) \times d}$.
The masked attention mechanism within the multimodal encoder plays a key role in achieving robustness under partial data availability. 
When one modality (or even more when more than two modalities are available) is missing, a binary modality mask is used to exclude its tokens from both the attention and gradient computations.

The latter, i.e., $\mathbf{m}_i^{\text{mm}} \in \{0,1\}^{\frac{d_v}{d} + F}$, is the composite mask derived from $m_i^{\text{img}}$ and $\mathbf{m}_i^{\text{tab}}$, where the value of the mask $m_i^{\text{img}}$ is repeated $\frac{d_v}{d}$ times equal to the number of visual tokens produced.
Similarly to before, we expand this masking vector into a matrix, and the encoder computes the following masked self-attention transformation: 
\begin{equation}\scalebox{.9}{$
\begin{aligned}
&\text{Attn}(\mathbf{q}_i^\text{mm}, \mathbf{k}_i^\text{mm}, \mathbf{v}_i^\text{mm}, \mathbf{m}_i^{\text{mm}}) = \\ &\text{ReLU}\left(\text{Softmax}\!\left(\frac{\mathbf{q}_i^\text{mm} (\mathbf{k}_i^\text{mm})^\text{T}}{\sqrt{d}} + \log(\mathbf{m}_i^{\text{mm}})\right) + \log(\mathbf{m}_i^{\text{mm}})^\text{T}\right)\mathbf{v}_i^\text{mm}.
\end{aligned}
$}
\end{equation}
Therefore, the multimodal representation $\mathbf{z}_i$ produced by the multimodal encoder integrates only the available modalities and features, guaranteeing that missing representations do not influence the context of available ones. 
The final prediction is obtained through a classification head $f_{\theta_{cl}}$ fed with the flattened version of the multimodal representation $\mathbf{z}_i$:
\begin{equation}
\mathbf{\hat{y}}_i = f_{\theta_{cl}}(\mathbf{z}_i).
\end{equation}


\subsection{Modality-Dropout Regularization}\label{sec:modality_dropout}

To further enhance resilience under missing-modality conditions, we introduce a regularization strategy that implements a \textit{modality-dropout} approach. 
It acts in synergy with the masked attention mechanisms introduced in sections~\ref{sec:tabular_encoder} and \ref{sec:multimodal_encoder}, which handles deterministic missingness originating from the dataset, modality-dropout introduces controlled stochastic missingness during training as a form of regularization. 
This complementarity enables the model to develop adaptive fusion strategies that remain effective across a continuum of data completeness conditions.

Inspired by classical dropout, our approach operates at the modality level rather than on individual neurons or layers.  
Indeed, given a patient $i$ where $m_i^\text{img} + \text{max}(\mathbf{m}_i^\text{tab})>1$, a Bernoulli variable $r_i$ 
determines whether a stochastic masking event occurs.
On the one hand, if the patient is selected for masking, one modality is randomly dropped among the two available, setting its mask to zero, i.e., 
$m_i^\text{img} = 0$ or $ \mathbf{m}_i^\text{tab}= \mathbf{0}$.
On the other hand, if a patient has only one modality in the dataset, modality dropout is disabled, preventing the complete removal of information.

It is worth noting that such a regularization offers two key advantages.
First, it prevents the model from over-relying on a single dominant modality, encouraging balanced feature learning and redundancy across sources when the patient includes all the available modalities. 
Second, it exposes the multimodal encoder to a wide range of missingness patterns during training, allowing it to learn to exploit the complementary information available from any subset of modalities. 
This stochastic exposure improves the model's ability to generalize at inference time, even under extreme missingness scenarios where one or more modalities are entirely absent.


\subsection{Training Objective}\label{sec:loss}
For all training phases, i.e., the unimodal pre-training and the multimodal fine-tuning that will be presented in the next section, we optimize the model using a multilabel binary cross-entropy loss computed only on the set of observed labels for each patient. 
For each class $c \in \{1,\dots,C\}$, let $m^{y}_{i,c} \in \{0,1\}$ denote whether the corresponding label $y_{i,c}$ of patient $i$ is observed. 
The training objective is a masked binary cross-entropy loss:
\begin{equation}\scalebox{0.85}{$
 \mathcal{L} = - \frac{1}{\sum_{i,c} m^{y}_{i,c}} 
 \sum_{i=1}^{N} \sum_{c=1}^{C}
 m^{y}_{i,c} \Big[
 y_{i,c}\log \hat{y}_{i,c} + (1-y_{i,c})\log (1-\hat{y}_{i,c})
 \Big].
$}\end{equation}
This label-wise masking ensures that the model receives consistent supervision even when the target vector is partially missing, as is common in clinical multilabel settings.


\subsection{Pre-training and Fine-tuning}\label{sec:training}
The proposed framework is trained end-to-end following a two-stage procedure designed to stabilize optimization and encourage balanced learning across modalities. 
Since the vision and tabular encoders operate on heterogeneous data types and exhibit different convergence behaviours, we first optimize each unimodal branch independently before performing multimodal fine-tuning.

\paragraph{Unimodal pre-training}
The vision encoder $f_{\theta_{vi}}$ and the tabular encoder $f_{\theta_{ta}}$ are initially trained on their respective modalities using the same cross-validation folds, defined in section~\ref{sec:cross_validation}, used in the next multimodal fine-tuning.
Indeed, we defined a unique stratified cross-validation scheme~\cite{bib:multilabel_strat}, and the corresponding folds are used both during unimodal and multimodal training to avoid any form of data leakage. 
A reduce-on-plateau scheduler reduces the learning rate when the validation loss plateaus, and an early-stopping criterion with a greater patience monitors training. 

\paragraph{Multimodal fine-tuning}
After pre-training, the multimodal encoder $f_{\theta_{mm}}$ and the classification head $f_{\theta_{cl}}$ are trained jointly with the unimodal branches. 
To prevent catastrophic forgetting of the pre-trained representations, while allowing the fusion mechanism to learn the joint representations, we update the parameter of the unimodal encoders with a smaller learning rate, corresponding to the order of magnitude reached at the end of the unimodal pre-training, while for the multimodal encoder we use a learning rate policy equal to the one used in the unimodal pre-training.

\section{Material}\label{sec:material}


We use a multimodal dataset obtained by pairing chest X-Rays from the MIMIC-CXR repository with structured clinical information from MIMIC-IV.
The multimodal dataset employed, resulting from the data preparation and filtering process reported in \ref{app:data}, contains 62071 imaging-clinical pairs, each including 1 scan and 10 clinical features (\tablename~\ref{tab:vital_signs}). 
Additional preprocessing was applied to the clinical features before training the multimodal models, which is described in section~\ref{sec:preprocessing}. 

Label information consisted of 14 diagnostic findings extracted for each study. 
Owing to substantial label incompleteness, both training and evaluation employed label-wise masking by $m^{y}_{i,c}$ to ensure that only observed labels contributed to the optimization objective and performance metrics. 
This design reflects common clinical documentation practices and provides a realistic testing environment for evaluating multimodal learning under heterogeneous and irregular missingness patterns.
Table~\ref{tab:labels} summarizes the distribution of positive, negative, and missing labels across the dataset. 
The high missing rates of the annotations further motivates the need for models capable of robust inference under uncertain and partially observed multimodal conditions.

\begin{table}[ht]
 \centering
 \resizebox{\columnwidth}{!}{
 \begin{tabular}{c|c|c|c}
 \toprule
 \textbf{Class} & \textbf{\# of Negatives} & \textbf{\# of Positives} & \textbf{\# of Missing} \\
 \midrule
 Atelectasis & 33747 (54.37\%) & 6371 (10.26\%) & 21953 (35.37\%) \\
 Cardiomegaly & 34124 (54.98\%) & 5435 (8.75\%) & 22512 (36.27\%) \\
 Consolidation & 35743 (57.58\%) & 1110 (1.79\%) & 25218 (40.63\%) \\
 Edema & 35516 (57.22\%) & 4225 (6.81\%) & 22330 (35.97\%) \\
 Enlarged Cardiomediastinum & 33826 (54.49\%) & 413 (0.67\%) & 27832 (44.84\%) \\
 Fracture & 33763 (54.39\%) & 958 (1.54\%) & 27350 (44.07\%) \\
 Lung Lesion & 33725 (54.33\%) & 1278 (2.06\%) & 27068 (43.61\%) \\
 Lung Opacity & 33909 (54.63\%) & 8994 (14.49\%) & 19168 (30.88\%) \\
 No Finding & 24459 (39.40\%) & 33665 (54.24\%) & 3947 (6.36\%) \\
 Pleural Effusion & 35004 (56.39\%) & 5887 (9.49\%) & 21180 (34.12\%) \\
 Pleural Other & 33671 (54.24\%) & 240 (0.39\%) & 28160 (45.37\%) \\
 Pneumonia & 35903 (57.84\%) & 3262 (5.26\%) & 22906 (36.90\%) \\
 Pneumothorax & 35173 (56.67\%) & 623 (1.00\%) & 26275 (42.33\%) \\
 Support Devices & 32075 (51.67\%) & 4044 (6.52\%) & 25952 (41.81\%) \\
 \bottomrule
 \end{tabular}}
 \caption{List of available labels together with the numbers of their negative, positive and missing patients and rates in brackets.}
 \label{tab:labels}
\end{table}

\subsection{Data Preprocessing}\label{sec:preprocessing}

Before training the unimodal and multimodal models, both modalities underwent modality-specific preprocessing to ensure consistent and stable input representations. 
Chest X-Rays were resized to $224\times 224$ pixels, normalized to the $[0,1]$ range, and replicated across three channels to match the input format required by the ResNet-50 backbone. 
During training, we apply data augmentation with a 50\% probability and using random horizontal flipping and random rotations of $\pm15^\circ$. 

Clinical features were preprocessed using a deterministic preprocessing procedure, whose normalization parameters and category mappings were estimated on the training data and across all evaluated settings applied to the corresponding validation and test sets.
Numerical variables were scaled to the $[0,1]$ interval to harmonize their magnitude across different measurement units. 
Categorical variables (sex and ethnicity) were converted into integer indices corresponding to their discrete categories. 
The high-cardinality textual description of the presenting problem was mapped into a multi-hot encoding that retained only the categories observed in the dataset. 
After these transformations, the final structured clinical representation consisted of 381 features, each subsequently embedded as described in section~\ref{sec:tabular_encoder}.


\section{Competitors}\label{sec:competitors}

We compare our architecture against three representative strategies, such baselines correspond to the main methodological families outlined in the introduction, namely modality augmentation, feature space engineering, and model selection approaches. 
Since vision-tabular learning under arbitrary modality missingness remains an emerging setting, the literature does not yet provide a clearly established benchmark method, although several related approaches are available. 
For this reason, we focus on representative competitors that capture the main conceptual strategies used to handle missing modalities, rather than attempting an exhaustive comparison with all existing variants. 
To ensure a fair analysis, all competitor methods were built using the same unimodal encoders adopted in our framework described in sections~\ref{sec:vision_encoder} and \ref{sec:tabular_encoder}, whereas the multimodal module consisted of a multilayer perceptron (MLP) further described in section~\ref{sec:experimental_configuration}. 
This design choice maintains architectural consistency across methods, ensuring that differences in performance arise from the missing-modality strategy rather than from discrepancies in model capacity.

\paragraph{Modality Augmentation}
In the first category, we include the \textit{constant value modality composition} baseline, a simple yet widely adopted strategy for handling missing inputs~\cite{bib:modality_augmentation}. 
Under this approach, any missing representation, either from the vision or the tabular encoders, is replaced with an array of zeros.
This method preserves architectural completeness, enabling the model to operate without explicit missingness modeling. 
However, it treats missingness as if it were a meaningful numeric value, which can bias representation learning and introduce unintended distributional shifts. 

\paragraph{Feature Space Engineering}
This approach attempts to achieve resilience through deterministic fusion rules rather than architectural modifications: 
each modality is encoded separately using its unimodal encoder, and the final joint representation is obtained by max pooling across the available unimodal embeddings~\cite{bib:feature_space_engineering}.
In practice, the visual and tabular latent vectors are computed independently; if a modality is missing, its embedding is simply omitted from the pooling operation. 
Max pooling selects, dimension-wise, the most dominant activation among the present modalities, effectively functioning as a hard selection mechanism.

\paragraph{Model Selection}
This approach trains both unimodal and multimodal predictors using only patients with complete data modalities, and then, at inference time, it dynamically selects the appropriate one based on modality availability~\cite{bib:model_selection}. 
Under this strategy: (i) when only the CXR is available, predictions are generated by the pre-trained vision model; (ii) when only clinical features are available, the tabular model is used instead; (iii) when both modalities are present, a multimodal network, trained exclusively on full-data samples, provides the prediction.
This approach avoids the pitfalls of imputing or generating missing data and does not require specific fusion strategies able to operate on incomplete inputs. 


\subsection{Ablation analysis}\label{sec:ablation}
To disentangle the contribution of each component of the proposed framework and assess the effectiveness of its architectural choices, we conducted an ablation study organized along two complementary axes: (i) resilience mechanisms inherited from competitor strategies, (ii) and fusion strategy.

We first examined whether two representative families of missing-modality methods, specifically modality augmentation and model selection, could enhance or replace the masking-based design of our architecture when incorporated into our vision-tabular framework.
In the modality augmentation condition, the masking pipeline was disabled and any missing latent representation was replaced using the constant-value imputation rules described for the baseline in section~\ref{sec:competitors}. 
This allows us to test whether the improvements brought by our method stem from the explicit attention-level masking or whether a simpler strategy is sufficient.
In the model selection ablation, the multimodal model is trained on the fully available patients only, and then the model employed at inference time dynamically switches between the unimodal and multimodal approaches based on the modalities available, mirroring the logic of the corresponding baseline. 
This ablation isolates the benefit of integrating partial-modality inputs within a single model from that of relying on specialized unimodal predictors.
Although feature space engineering was included as a competitor, this family could not be meaningfully adapted to our architecture. 
Its fusion principle, pooling unimodal embeddings into a single latent vector, is incompatible with the attention-based encoder, which operates on token sequences rather than a single latent representation. 
Since this would require replacing the core mechanism that defines our model, the feature space engineering strategy was not included in the ablation variants.

To quantify the importance of intermediate fusion within our architecture, we further compared our approach against its early and late fusion counterparts.
In the early fusion variant, the unimodal encoders were kept frozen while training the multimodal component, preventing joint optimization across unimodal modules and isolating the contribution of the fusion layer alone.
In the late fusion variant, the unimodal outputs were combined through simple averaging, bypassing any cross-modal interaction within a shared transformer space.
These ablations evaluate whether the benefits of our model derive from the unified masked-attention fusion layer or simply from the presence of separate unimodal encoders.

Together, these ablations provide a structured analysis of how resilience to missing data emerges in our framework, whether from explicit masking, from the fusion mechanism, or from the interplay between modality-aware encoders and the multimodal transformer.


\section{Experiments}
\label{sec:experiments}

To comprehensively assess the resilience of the proposed multimodal architecture, we design two complementary experimental protocols, each targeting a distinct and clinically motivated research question.

RQ1: How resilient is the proposed framework when trained on increasingly incomplete multimodal data? 
In retrospective and multi-center clinical datasets, different modalities are often collected asynchronously or only for a subset of patients, resulting in highly unpaired training data. 
To address this question, we progressively increase the missing rate of one modality at a time during training, while keeping the other modality fully available. 
Missingness is introduced by randomly masking the selected modality for a given fraction of training patients, whereas the test set always contains fully available multimodal inputs. 
We consider missingness in training levels of 0\%, 25\%, 50\%, and 75\% for each modality. 
The maximum missing rate is intentionally capped at 75\% to ensure that a sufficient number of patients remain available to train the corresponding unimodal encoder, while also exposing the multimodal encoder to enough paired data to learn meaningful cross-modal interactions. 
This protocol evaluates whether the model can learn stable unimodal and multimodal representations even when a large fraction of the training data lacks one modality, and whether modality-dropout and masked attention prevent catastrophic degradation under severe but realistic training sparsity.

RQ2: How resilient is the proposed framework when deployed under partial or fully unimodal data availability? 
In routine clinical practice, one modality may be partially or entirely unavailable for a given patient due to acquisition delays, resource constraints, or incomplete records. 
To address this question, all models are trained on fully available multimodal data, while missingness is progressively introduced only at test time. 
For each modality, we evaluate performance under missingness levels of 0\%, 25\%, 50\%, 75\%, and 100\%, masking one modality at a time and leaving the other intact. 
Unlike the missingness in training scenario, missingness in test can reach 100\%, corresponding to a fully unimodal inference setting. 
This allows us to assess how the multimodal model's performance degrades as information is removed, and also directly compare unimodal performance before and after multimodal fine-tuning. 
This protocol captures realistic operational scenarios such as missing imaging studies, delayed acquisitions, or unavailable clinical records, and evaluates whether the multimodal encoder can reliably condition its predictions exclusively on the observed information.

In both protocols, a random subset of patients had one of the available modalities removed by independently applying the missing rate to the imaging and tabular domains, thus producing two parallel stress curves:
\begin{itemize}
\item Tabular-missingness curve: progressively masking the clinical variables while keeping the radiographs intact;
\item Imaging-missingness curve: progressively masking the CXR modality while keeping the clinical features intact.
\end{itemize}
This design isolates the contribution of each modality and allows us to quantify how strongly the model depends on imaging versus tabular information in different prediction tasks.

\subsection{Experimental Configuration}\label{sec:experimental_configuration}

All models were trained under a unified optimization protocol to ensure consistency across unimodal and multimodal configurations.
Training was performed with a batch size of 512, for a maximum of 500 epochs, using the AdamW optimizer and an initial learning rate of 0.001.
A 50-epoch warm-up phase was employed during which neither the scheduler nor the early-stopping patience counters were updated, allowing the models to stabilize before adaptive learning rate adjustments.
After warm-up, a Reduce-on-Plateau scheduler reduced the learning rate by a factor of 10 whenever the validation loss failed to improve for 25 consecutive epochs; early stopping was governed by the same validation-loss criterion but had 50 epochs of patience.

The unimodal encoders were trained independently and served as the initialization for the multimodal model. 
The vision encoder consisted of a ResNet-50, producing a 2048-dimensional latent vector.
The tabular encoder was implemented as a transformer with 2 encoder layers, 16 attention heads, and 1024-dimensional tokens. 
All unimodal models were trained on 2 NVIDIA A40 GPUs.

Following unimodal convergence, the multimodal model was trained end-to-end by concatenating the unimodal representations into a joint sequence by reshaping the vision token into two tokens of size 1024 to match the dimension of the tabular token.
The multimodal encoder mirrored the tabular architecture, employing 2 transformer layers and 16 attention heads. 
Its parameters were optimized using the same learning rate policy adopted for unimodal training, while the parameters of the vision and tabular encoders were fine-tuned with a learning rate three orders of magnitude smaller, preventing catastrophic drift from their pre-trained representations. 
To improve resilience to partial data availability, a 30\% probability was set to the $r_i$ variable in the modality-dropout regularization during multimodal training. 
In the competitors’ models, the multimodal encoder was implemented as an MLP composed of three layers, each with 300 neurons and ReLU activations.
Multimodal training required 4 NVIDIA A40 GPUs due to the increased memory demands associated with joint token processing.

\paragraph{Stratification Strategy}\label{sec:cross_validation}
We adopted a five-fold stratified cross-validation scheme for the test sets.
Within each training fold, the validation set was obtained through a 20/80 stratified holdout.
Stratification followed the method presented in~\cite{bib:multilabel_strat}, which extends classical stratification to multilabel datasets by preserving not only marginal label frequencies but also the distribution of label co-occurrences across folds.
This ensures that rare combinations of diagnostic findings are across all evaluated settings represented in all partitions, thereby reducing sampling bias and improving the reliability of comparative evaluation.

\paragraph{Evaluation Metric}
Performance was assessed using the weighted average AUC across the 14 diagnostic classes. 
For each fold, the AUC was computed per class and then aggregated using weights proportional to the ratio of positive patients for that class. 
This weighting scheme ensures that the final metric reflects the relative prevalence of each diagnostic category within the dataset. 
Final results are reported as the mean weighted AUC across the five cross-validation folds. 


\section{Results and Discussion}\label{sec:results}

To facilitate interpretation of the experimental findings, we first describe how to read the stress-test plots reported in \figurename s~\ref{fig:results}-\ref{fig:ablation_results_fusion}. 
Each panel contains a set of curves that quantify model degradation under controlled missingness conditions for one modality at a time, either during training or testing, separately. 
The tabular-missingness curves report performance while the clinical features are progressively masked ($0\%$-$75\%$ in training and $0\%$-$100\%$ in testing), keeping the imaging modality fully available. 
Conversely, the imaging-missingness curves measure performance as the X-rays are increasingly removed while clinical data remain intact. 
For reference, each plot shows a black dotted line corresponding to the unimodal performance of the relevant branch, enabling a direct comparison between multimodal fine-tuned models and their unimodal counterparts.
In the missingness in training experiments, the unimodal reference curve corresponds to the unimodal model associated with the modality whose availability is progressively reduced, providing a baseline for assessing performance degradation as multimodal supervision becomes increasingly unpaired.
In contrast, in the missingness in test experiments, the unimodal reference corresponds to the unimodal model of the modality that remains fully available throughout the experiment, i.e., the complementary modality with respect to the one being progressively masked. 
In this setting, the $100\%$ missingness point represents an interesting comparison configuration, corresponding to a purely unimodal inference of the multimodal approaches.
A detailed breakdown of the exact numerical values underlying all stress-test plots is provided in \ref{app:results}, which includes full performance tables across all missingness conditions, enabling precise inspection of each model’s behavior.

\subsection{RQ1: Resilience under Missingness in Training}\label{sec:rq1}
This section investigates how the proposed framework behaves when trained on increasingly incomplete multimodal data, a scenario that mirrors retrospective datasets where modalities are collected asynchronously or only for selected subpopulations.

\subsubsection{Comparison with Competitors}
\begin{figure*}[!ht]
 \centering
 \resizebox{\textwidth}{!}{
 \begin{tikzpicture}
 \node[label={[xshift=.3cm, yshift=-.2cm]above: \textbf{Tabular-Missingness Curves}}] (A) at (0,0) {\includegraphics[width=.48\linewidth, trim=1.5cm 0 4.1cm 1.5cm, clip]{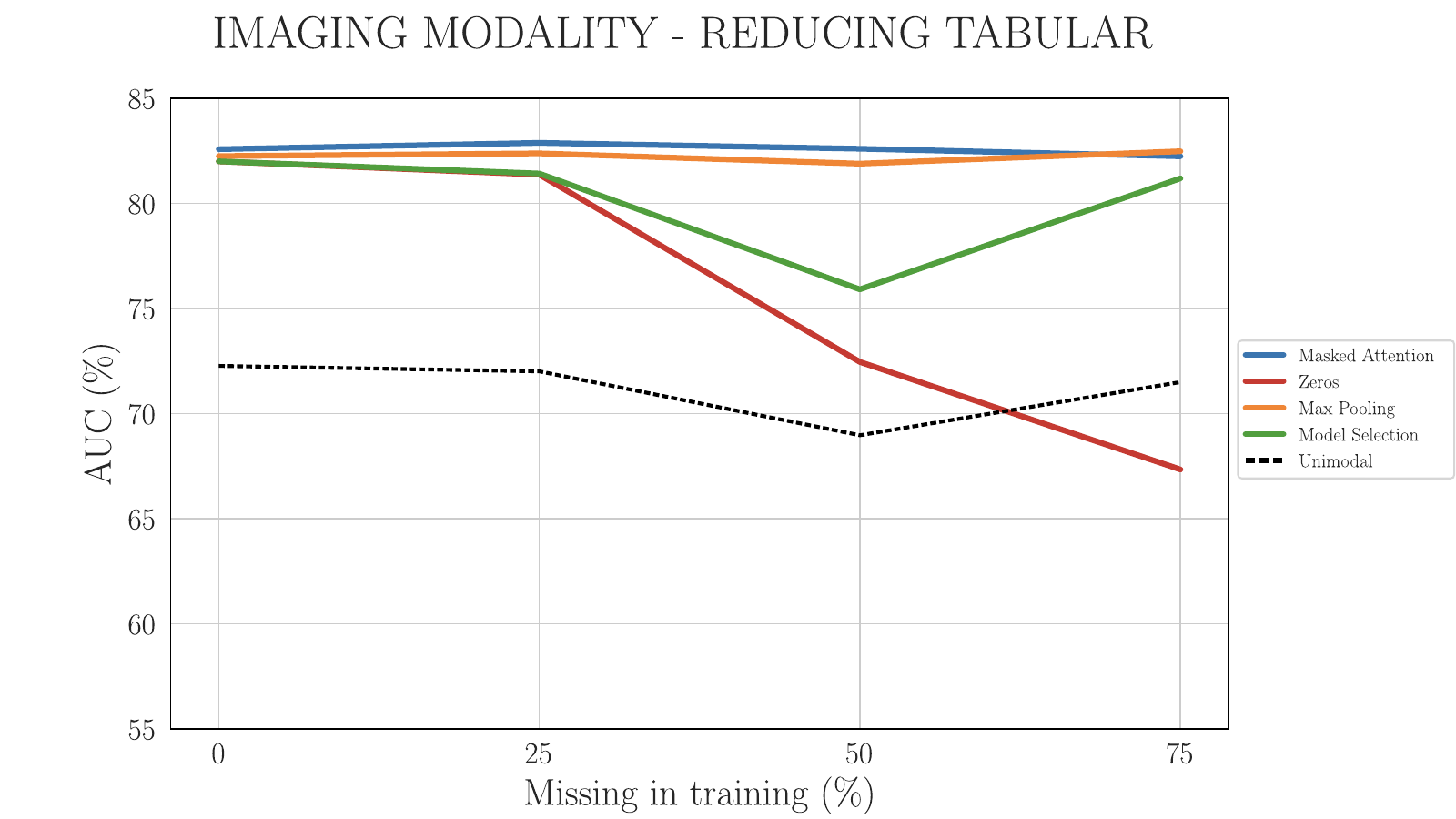}};
 \node[label={[xshift=.3cm, yshift=-.2cm]above: \textbf{Imaging-Missingness Curves}}] (B) [right=.2cm of A] {\includegraphics[width=.48\linewidth, trim=1.5cm 0 4.1cm 1.5cm, clip]{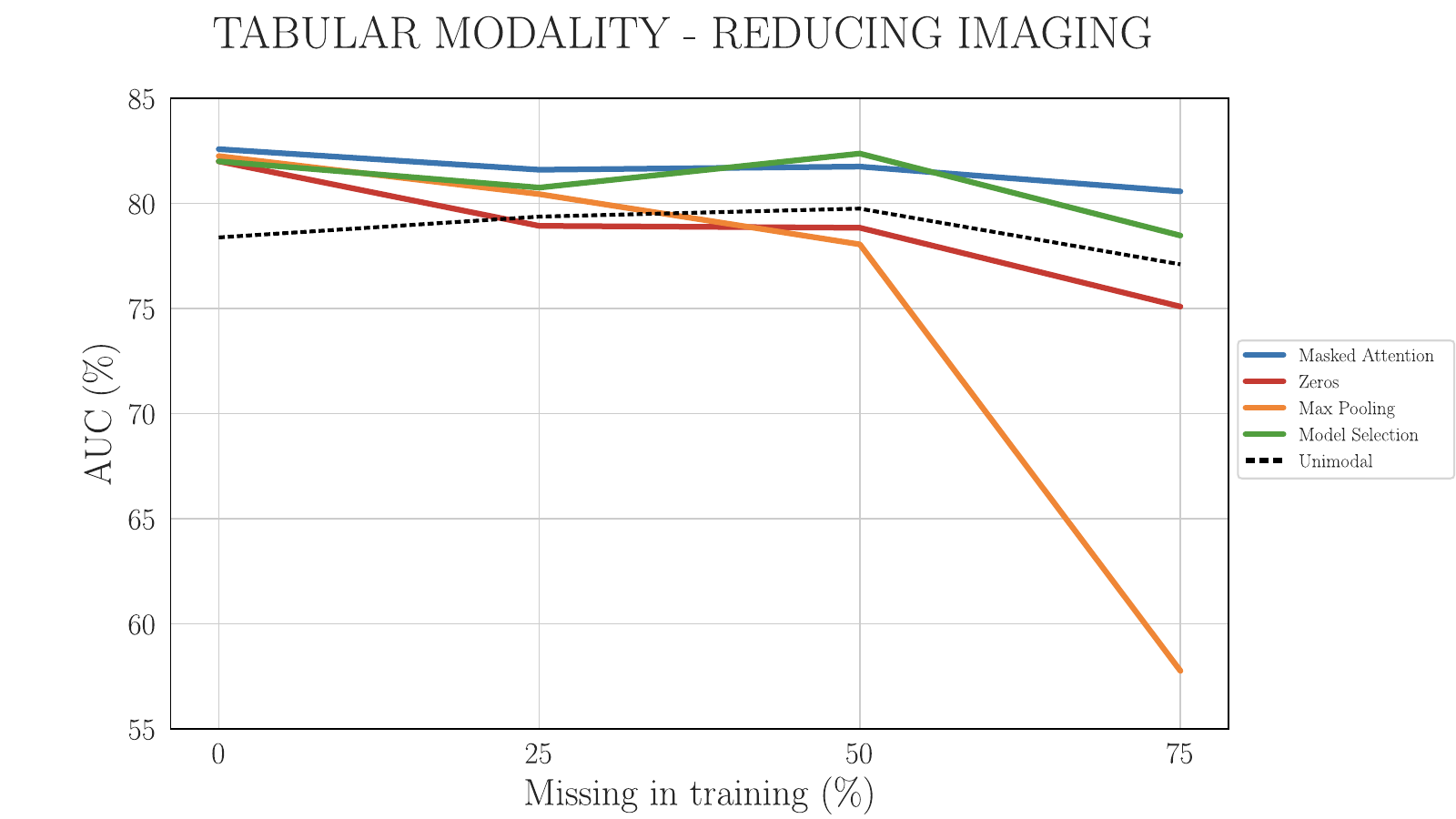}};
 \node (leg) at ($([yshift=-.3cm]A.south west)!0.5!([yshift=-.3cm]B.south east)$) {\includegraphics[height=.5cm]{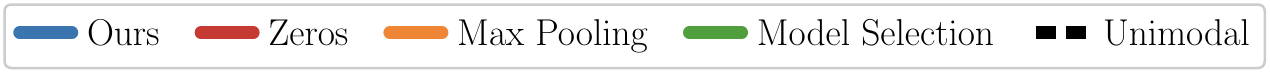}};
 \node (C) [below=.6cm of A] {\includegraphics[width=.48\linewidth, trim=1.5cm 0 4.1cm 1.5cm, clip]{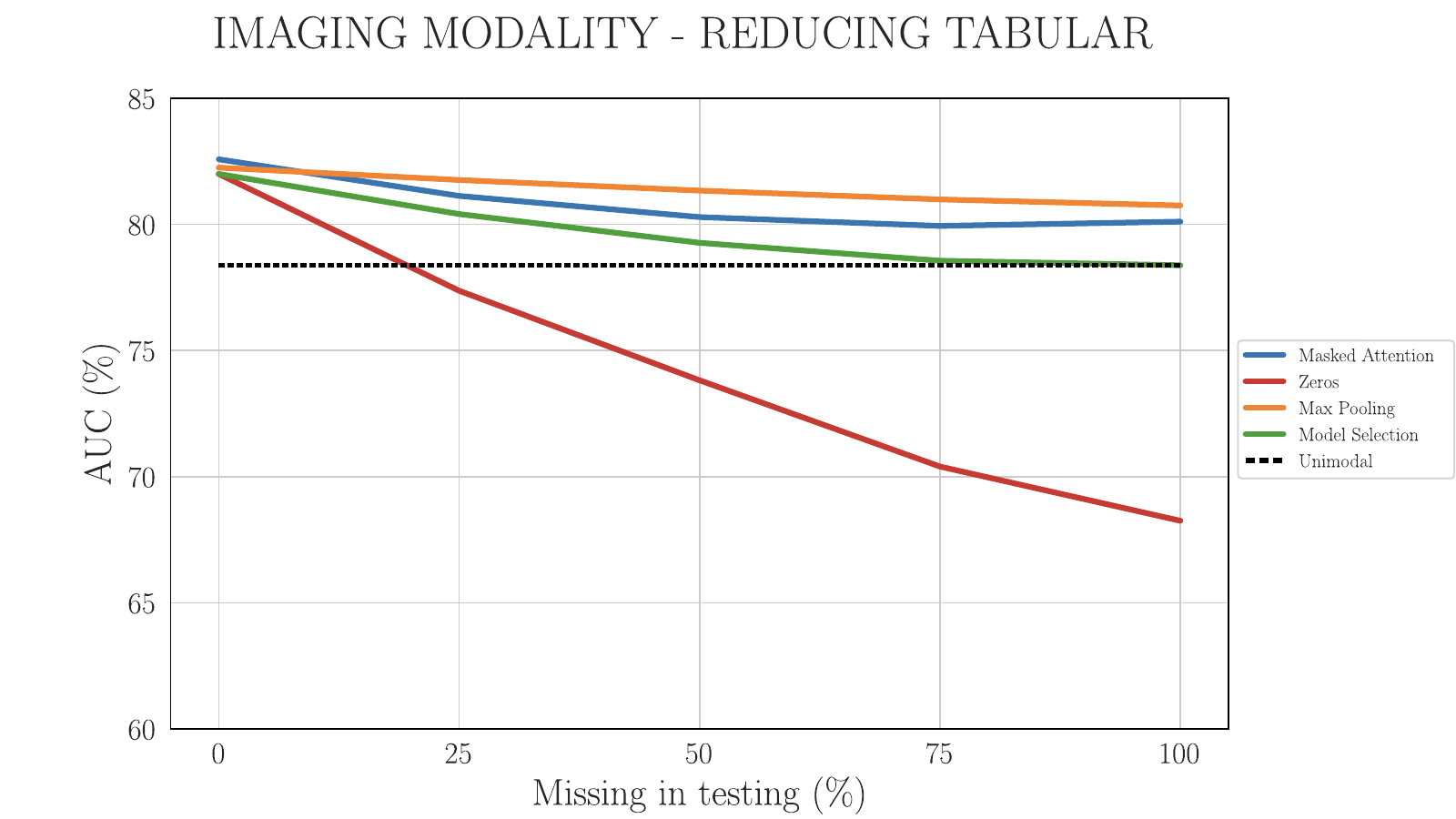}};
 \node (D) [right=.2cm of C] {\includegraphics[width=.48\linewidth, trim=1.5cm 0 4.1cm 1.5cm, clip]{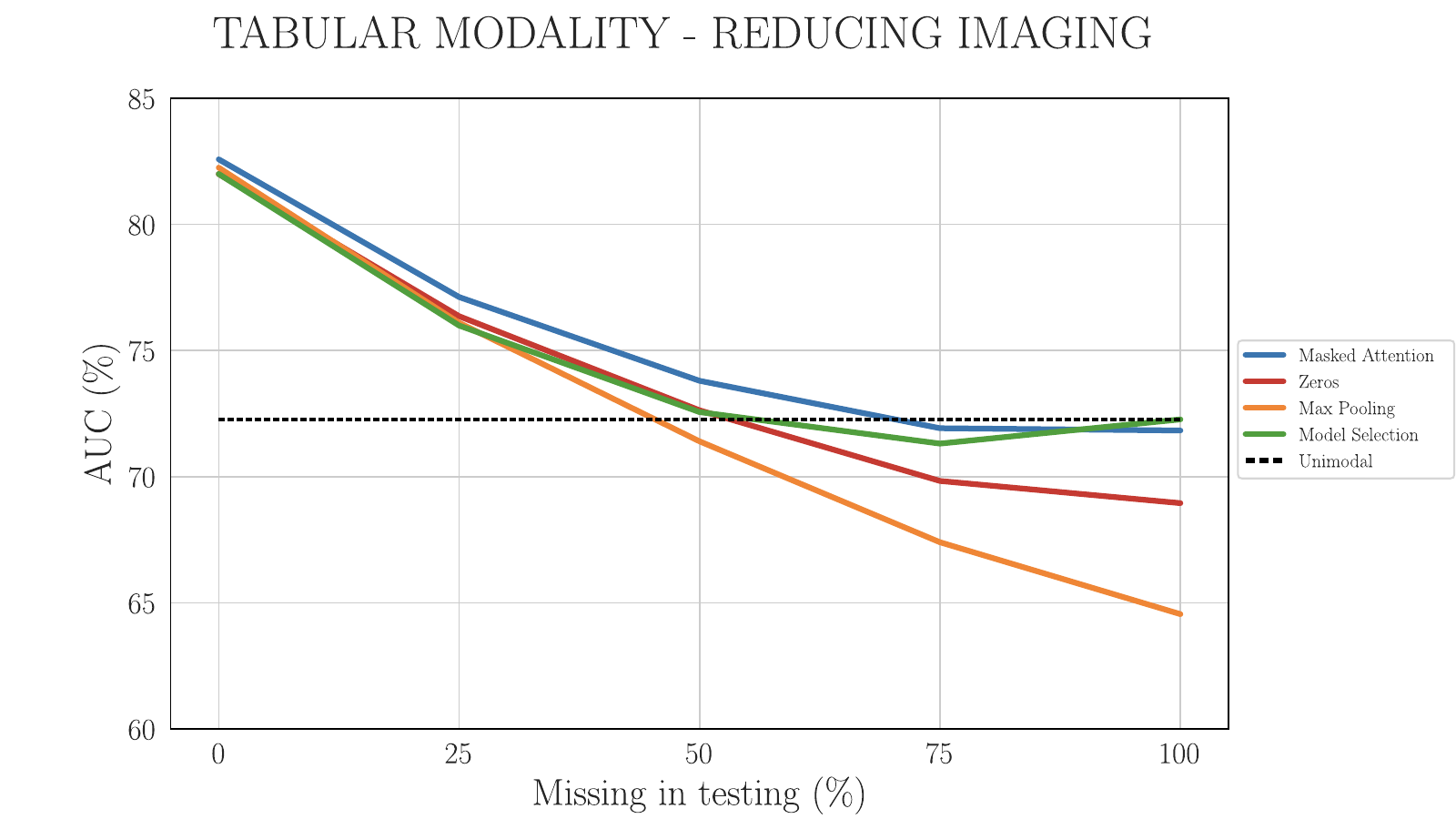}};
 \node[yshift=.25cm] (E) [left=0cm of A.west] {\rotatebox[origin=c]{90}{\textbf{Training Missingness}}};
 \node[yshift=.25cm] (F) [left=0cm of C.west] {\rotatebox[origin=c]{90}{\textbf{Test Missingness}}};
 \end{tikzpicture}}
 \caption{Stress-test curves (weighted AUC) of the proposed multimodal model and competing approaches under controlled modality missingness in training (top panels) and in test (bottom ones).
The left panels report performance when the tabular modality is progressively masked, while imaging data remain fully available. On the contrary, the right panels report performance when the imaging modality is progressively masked, keeping clinical features fully available. Curves correspond to the proposed masked-attention, the zeros modality composition, the max pooling and the model-selection approaches, all implemented using the same unimodal encoders and training protocol. The dotted black lines in the training missingness scenario denote the unimodal reference performance associated with the modality whose availability is reduced during training, providing a baseline for assessing degradation as multimodal supervision becomes increasingly unpaired. Conversely, horizontal dotted lines in the missingness in test scenario indicate the performance of the corresponding unimodal encoders evaluated independently on the fully available modality.}
 \label{fig:results}
\end{figure*}

The top panels of \figurename~\ref{fig:results} report the performance obtained when the missing rate of one modality is progressively increased during training, while the test set remains fully multimodal (see also \tablename~\ref{tab:results_train}). 
Across both tabular-missingness and imaging-missingness regimes, the proposed masked-attention model across all evaluated settings achieves the best performance and exhibits remarkably stable behavior, even when up to 75\% of the training data are unpaired. 
This indicates that explicit attention-level masking, combined with modality-dropout, allows the model to learn robust multimodal representations despite severe training data sparsity.

In contrast, the competing strategies display distinct undesirable behaviors depending on which modality is removed. 
The zeros modality composition baseline collapses when the tabular modality is progressively removed, indicating that replacing missing clinical features with constant values severely interferes with representation learning when the remaining supervision is dominated by high-dimensional visual inputs (\figurename~\ref{fig:results}, top left). 
Conversely, the max pooling strategy collapses when the imaging modality is increasingly removed, revealing a strong over-reliance on visual features and an inability to learn a competitive tabular representation when cross-modal supervision becomes sparse (\figurename~\ref{fig:results}, top right). 
The model-selection approach remains more stable than these heuristics but across all evaluated settings underperforms the proposed method, as it relies on fully paired data to train the multimodal branch and cannot exploit partially observed multimodal data during learning.

These results highlight that heuristic fusion or switching strategies are insufficient to cope with highly unpaired training data, whereas the proposed masked-attention framework maintains both convergence and performance across all missingness regimes.

\subsubsection{Ablation Analyses}

\begin{figure*}[!ht]
 \centering
 \resizebox{\textwidth}{!}{
 \begin{tikzpicture} 
 \node[label={[xshift=.3cm, yshift=-.2cm]above: \textbf{Tabular-Missingness Curves}}] (A) at (0,0) {\includegraphics[width=.48\linewidth, trim=1.5cm 0 4.1cm 1.5cm, clip]{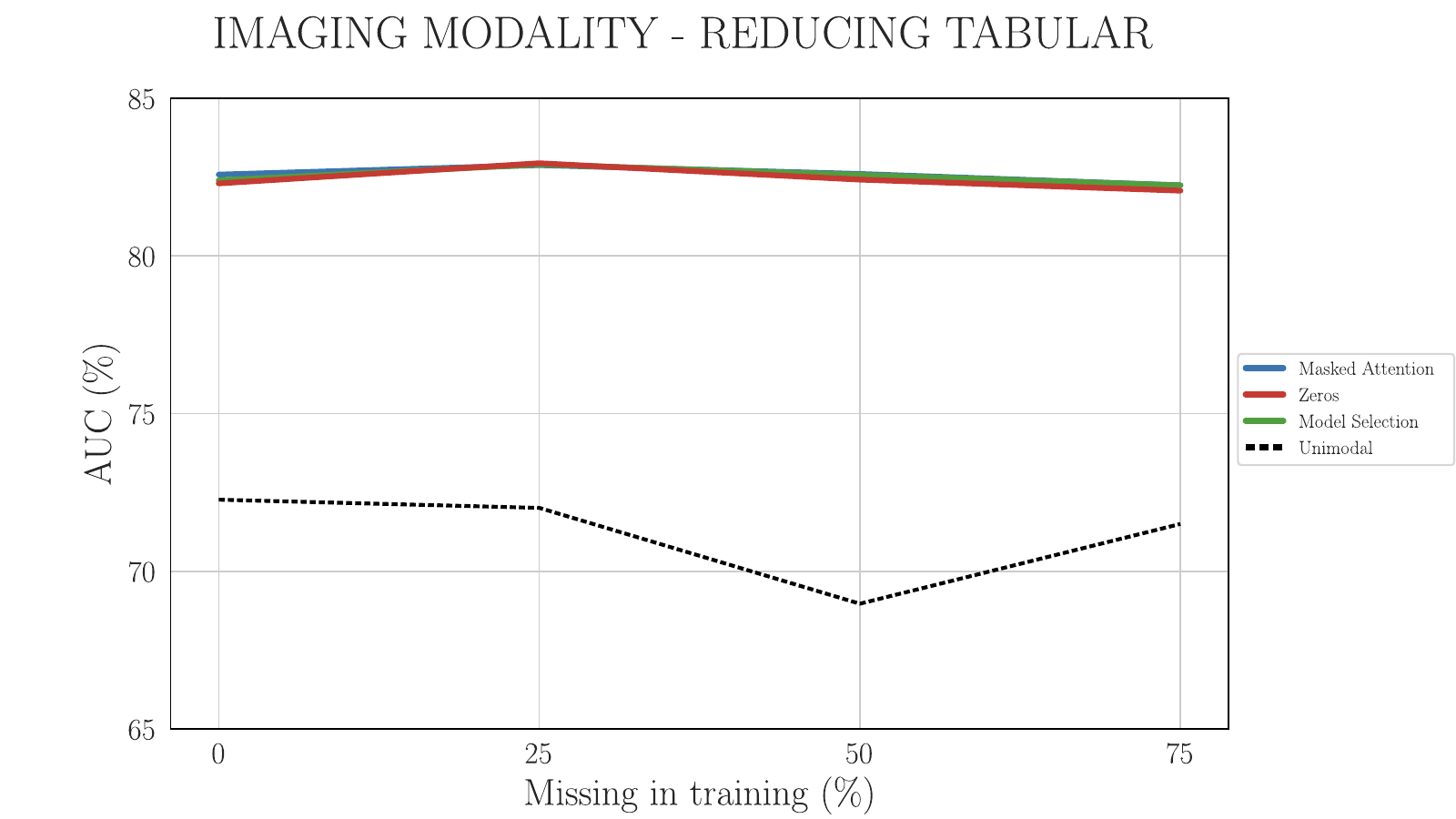}};
 \node[label={[xshift=.3cm, yshift=-.2cm]above: \textbf{Imaging-Missingness Curves}}] (B) [right=.2cm of A] {\includegraphics[width=.48\linewidth, trim=1.5cm 0 4.1cm 1.5cm, clip]{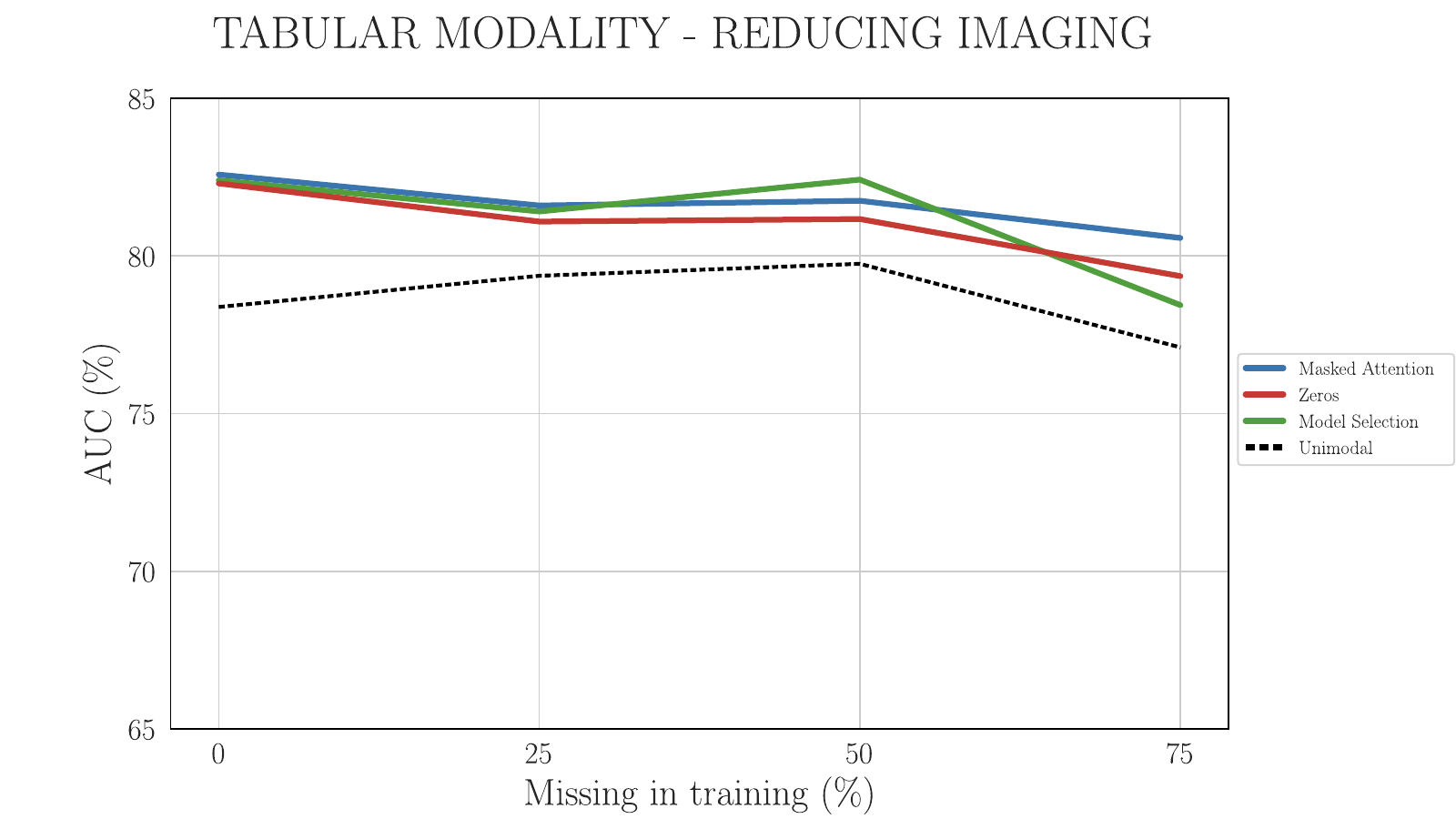}};
 \node (leg) at ($([yshift=-.3cm]A.south west)!0.5!([yshift=-.3cm]B.south east)$) {\includegraphics[height=.5cm]{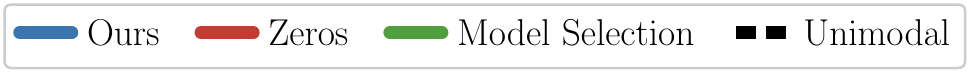}};
 \node (C) [below=.6cm of A] {\includegraphics[width=.48\linewidth, trim=1.5cm 0 4.1cm 1.5cm, clip]{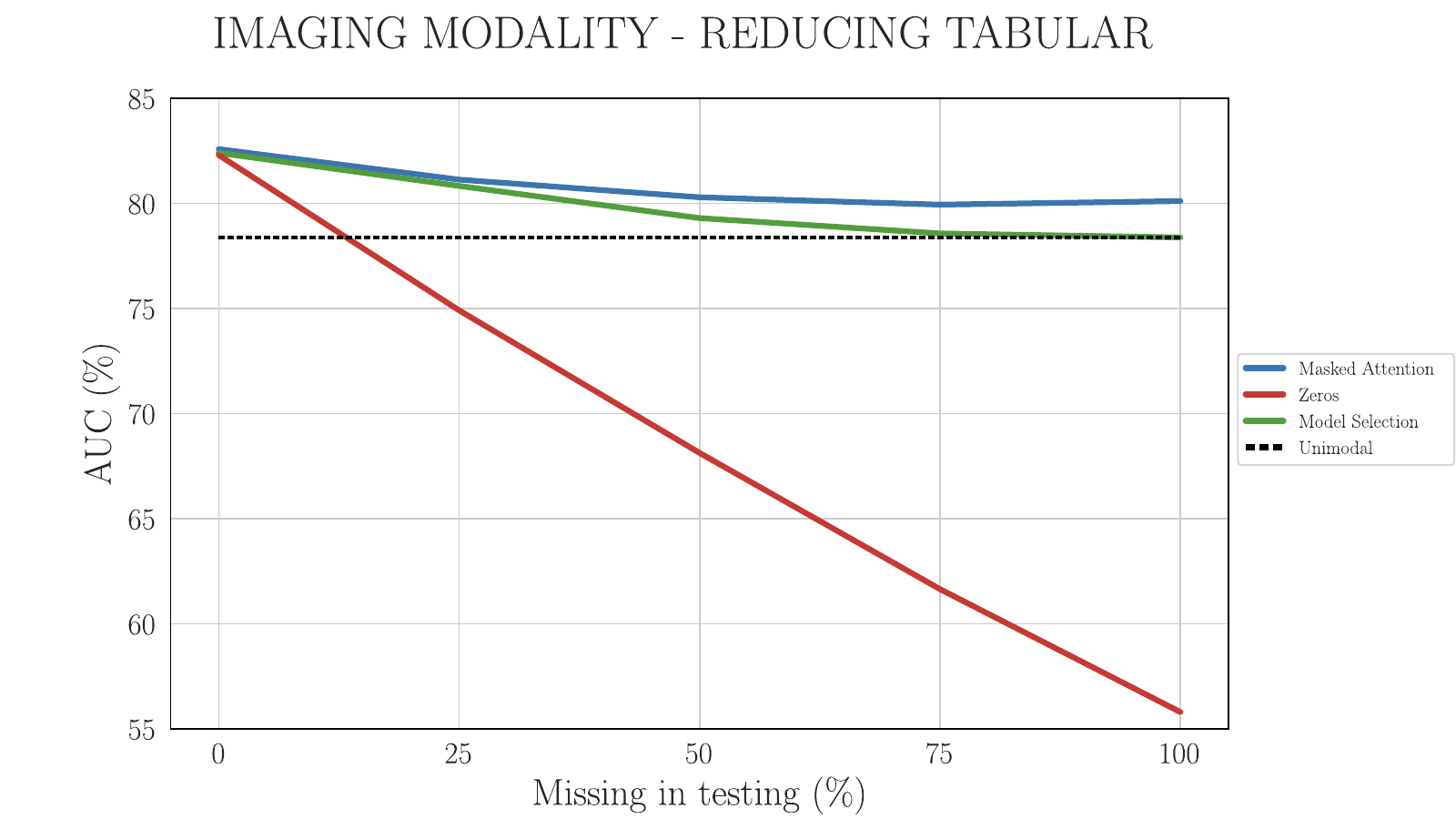}};
 \node (D) [right=.2cm of C] {\includegraphics[width=.48\linewidth, trim=1.5cm 0 4.1cm 1.5cm, clip]{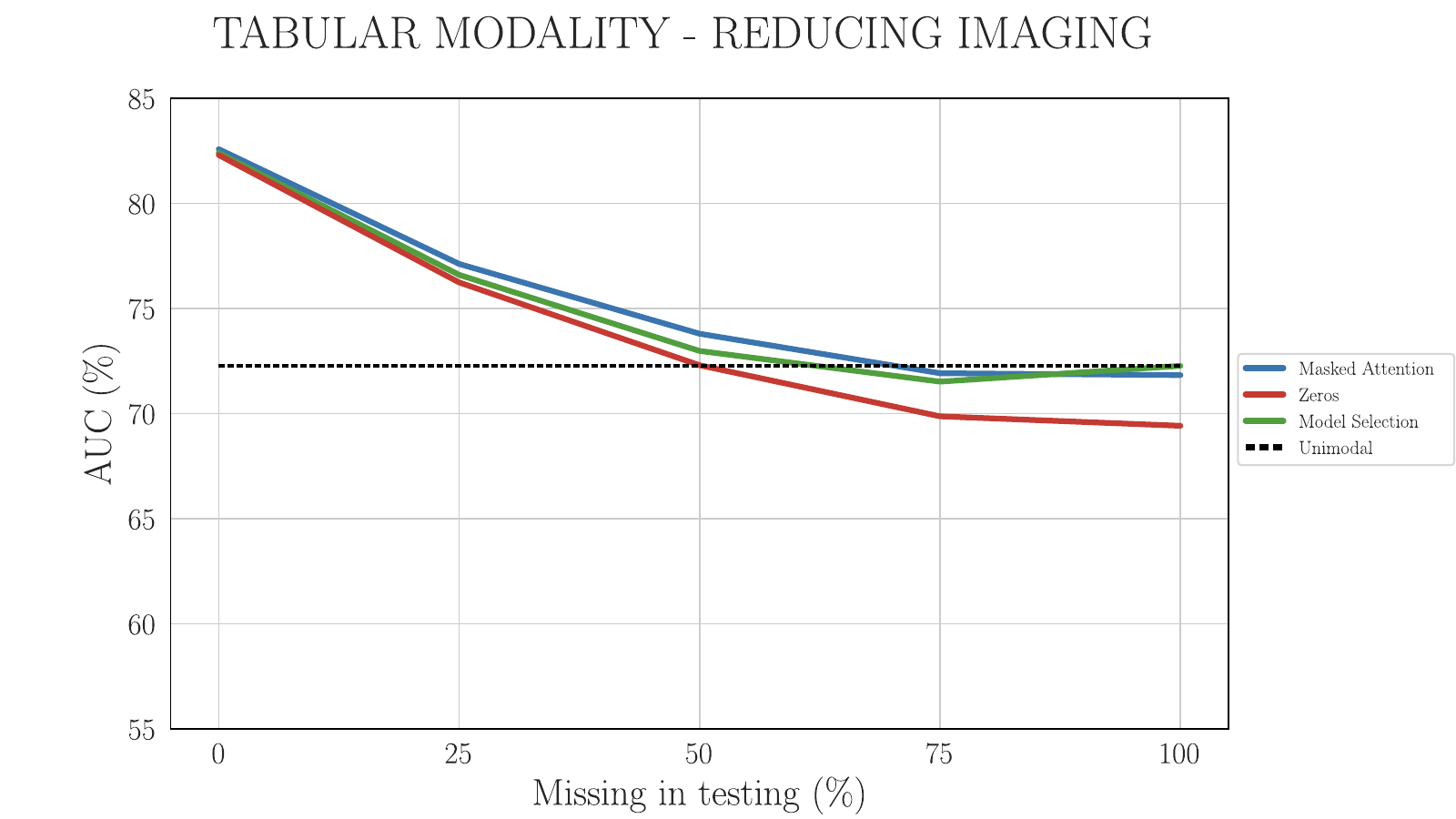}};
 \node[yshift=.25cm] (E) [left=0cm of A.west] {\rotatebox[origin=c]{90}{\textbf{Training Missingness}}};
 \node[yshift=.25cm] (F) [left=0cm of C.west] {\rotatebox[origin=c]{90}{\textbf{Test Missingness}}};
 \end{tikzpicture}}
 \caption{Stress-test curves (weighted AUC) obtained by applying competitor-inspired missing-modality strategies within the proposed architecture, while introducing missing modalities in training (top panels) and in test (bottom ones).
The left panels report performance when the tabular modality is progressively masked, while imaging data remain fully available. On the contrary, the right panels report performance when the imaging modality is progressively masked, keeping clinical features fully available. Curves correspond to the proposed masked-attention model, the zeros modality composition variant, and the model-selection variant, all implemented using the same unimodal encoders and training protocol. The dotted black lines in the training missingness scenario denote the unimodal reference performance associated with the modality whose availability is reduced during training, providing a baseline for assessing degradation as multimodal supervision becomes increasingly unpaired. Conversely, horizontal dotted lines in the missingness in test scenario indicate the performance of the corresponding unimodal encoders evaluated independently on the fully available modality.}
 \label{fig:ablation_results_approach}
\end{figure*}

\paragraph{Competitor-inspired Variants within Our Architecture} 
\figurename~\ref{fig:ablation_results_approach} (top panels) reports performance for the ablation configurations in which competitor strategies were integrated into our model architecture (see also \tablename~\ref{tab:ablation_results_train}). 
All variants exhibit comparable behavior when the tabular modality is progressively removed (\figurename~\ref{fig:ablation_results_approach}, top left). 
This limited sensitivity reflects both the inherently sparse nature of the clinical features and the dominant contribution of imaging during training, which mitigates the impact of additional tabular missingness. 
In contrast, clear differences emerge when increasing imaging missingness (\figurename~\ref{fig:ablation_results_approach}, top right). 
The zeros composition variant shows a pronounced performance degradation, indicating that constant-value replacement disrupts representation learning when high-dimensional visual information is absent. 
The model-selection variant, instead, is increasingly influenced by the unimodal baselines, reflecting its inability to exploit partially observed multimodal samples during training, as multimodal supervision is restricted to fully paired patients only.

\begin{figure*}[!ht]
 \centering
 \resizebox{\textwidth}{!}{
 \begin{tikzpicture} 
 \node[label={[xshift=.3cm, yshift=-.2cm]above: \textbf{Tabular-Missingness Curves}}] (A) at (0,0) {\includegraphics[width=.48\linewidth, trim=1.5cm 0 3.3cm 1.5cm, clip]{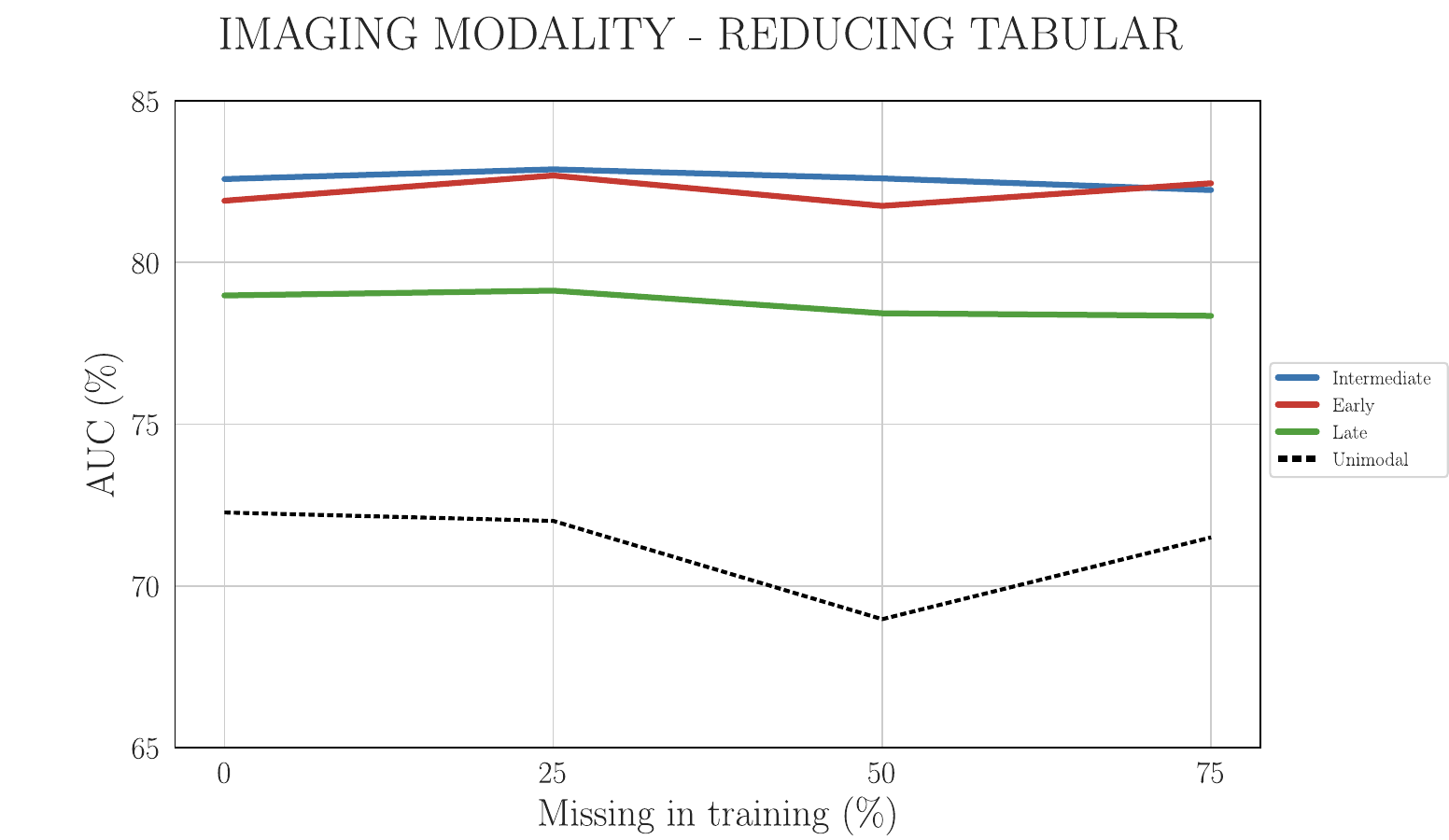}};
 \node[label={[xshift=.3cm, yshift=-.2cm]above: \textbf{Imaging-Missingness Curves}}] (B) [right=.2cm of A] {\includegraphics[width=.48\linewidth, trim=1.5cm 0 3.3cm 1.5cm, clip]{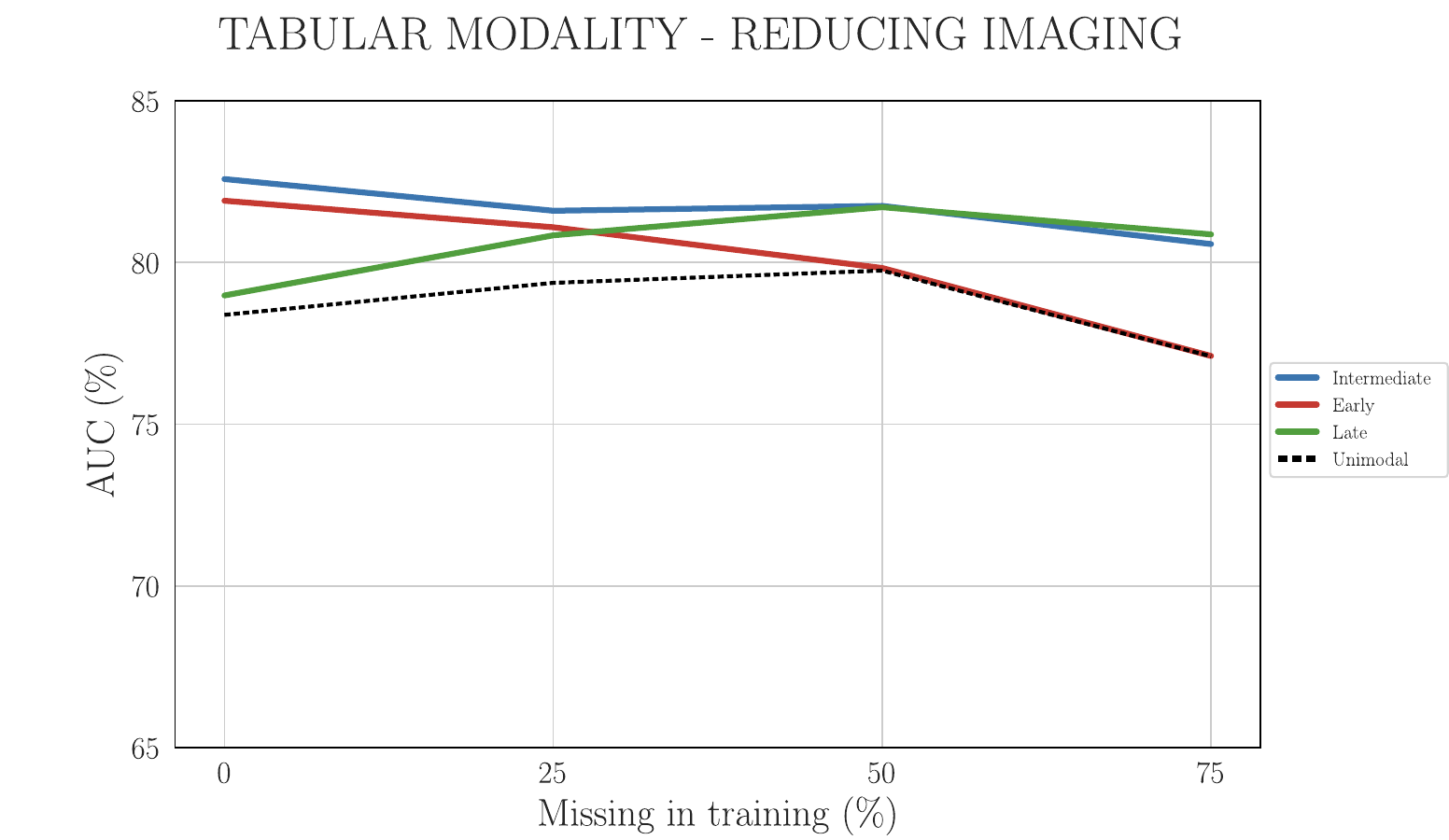}};
 \node (leg) at ($([yshift=-.3cm]A.south west)!0.5!([yshift=-.3cm]B.south east)$) {\includegraphics[height=.5cm]{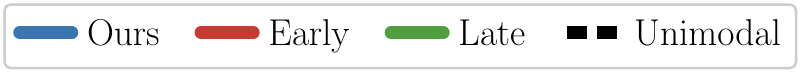}};
 \node (C) [below=.6cm of A] {\includegraphics[width=.48\linewidth, trim=1.5cm 0 3.3cm 1.5cm, clip]{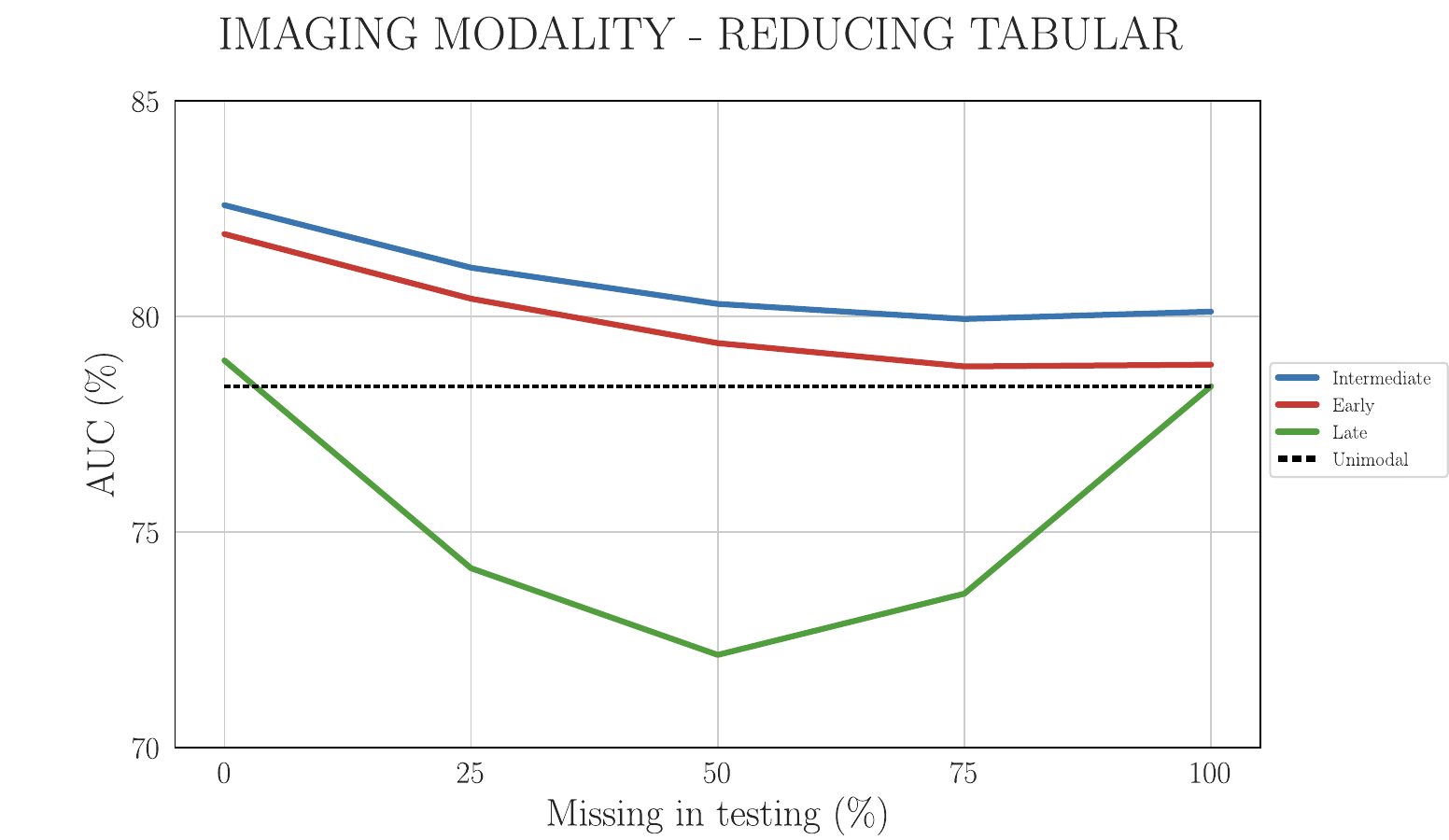}};
 \node (D) [right=.2cm of C] {\includegraphics[width=.48\linewidth, trim=1.5cm 0 3.3cm 1.5cm, clip]{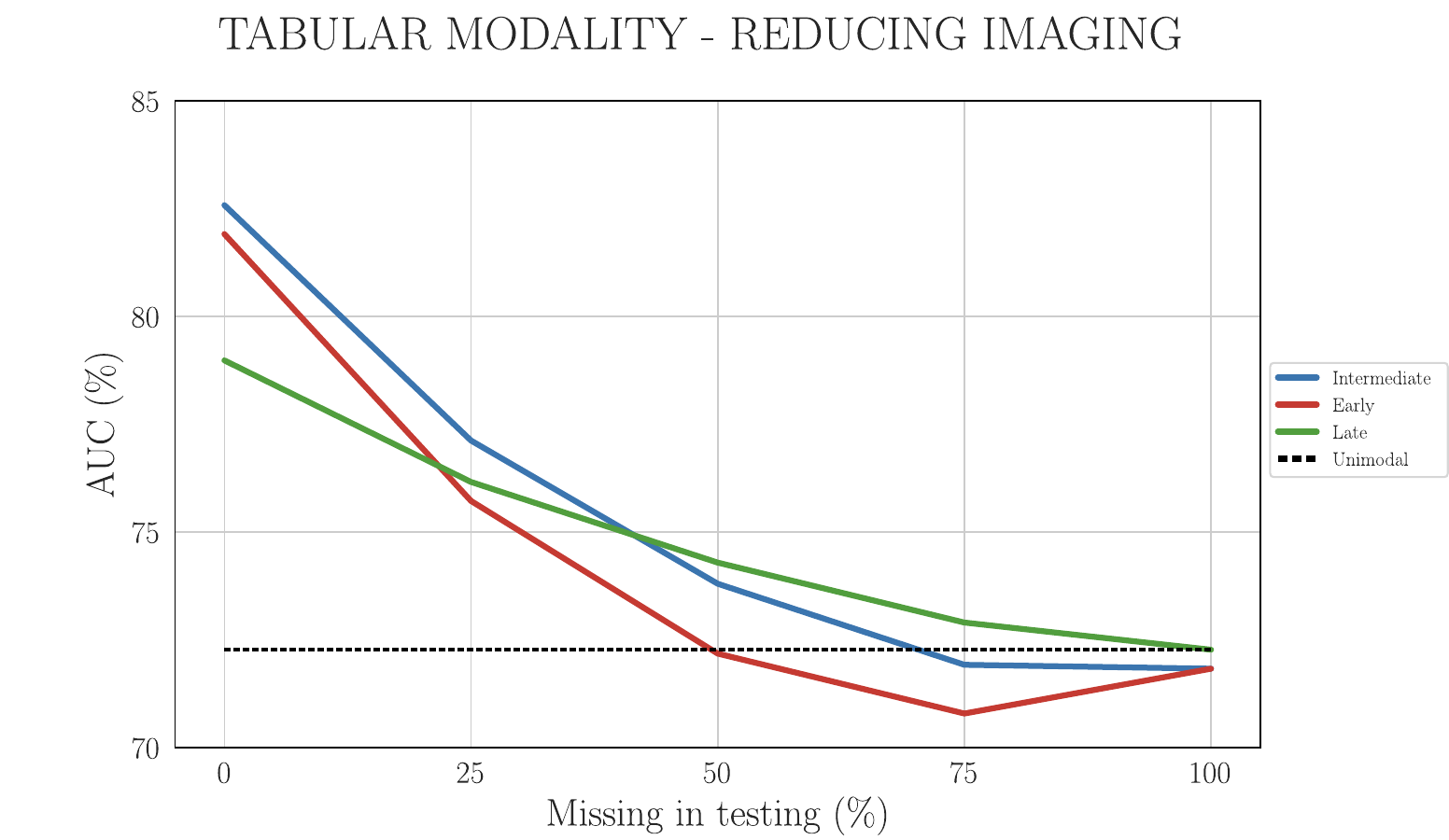}};
 \node[yshift=.25cm] (E) [left=0cm of A.west] {\rotatebox[origin=c]{90}{\textbf{Training Missingness}}};
 \node[yshift=.25cm] (F) [left=0cm of C.west] {\rotatebox[origin=c]{90}{\textbf{Test Missingness}}};
 \end{tikzpicture}}
 \caption{Performance curves (weighted AUC) comparing different fusion strategies under progressive modality removal in training (top panels) and in test (bottom ones).
The left panels report performance when the tabular modality is progressively masked, while imaging data remain fully available. On the contrary, the right panels report performance when the imaging modality is progressively masked, keeping clinical features fully available. Curves correspond to the proposed intermediate fusion model, and its early and late fusion counterparts, all implemented using the same unimodal encoders and training protocol. The dotted black line denotes the unimodal reference performance associated with the modality whose availability is reduced during training, providing a baseline for assessing degradation as multimodal supervision becomes increasingly unpaired. Conversely, horizontal dotted lines in the missingness in test scenario indicate the performance of the corresponding unimodal encoders evaluated independently on the fully available modality.}
 \label{fig:ablation_results_fusion}
\end{figure*}

\paragraph{Alternative Fusion Strategies} We further analyze the missingness in training scenario by comparing the proposed intermediate fusion transformer with its early and late fusion counterparts (\figurename~\ref{fig:ablation_results_fusion}, top panels; see also \tablename~\ref{tab:ablation_results_train}). 
Consistent with previous observations, removing the tabular modality does not markedly affect performance for any of the approaches (\figurename~\ref{fig:ablation_results_fusion}, top left). 
This behavior reflects the comparatively weaker predictive contribution of clinical variables in this setting and confirms that neither the proposed method nor its fusion variants are strongly dependent on tabular supervision during training. 
Nevertheless, subtle differences emerge in terms of stability. 
In particular, the early fusion variant exhibits slightly more unstable behavior across missingness levels, which can be attributed to the fact that unimodal encoders are kept frozen during multimodal training and therefore do not benefit from joint fine-tuning. 
In contrast, the late fusion strategy across all evaluated settings achieves lower performance, demonstrating its limited ability to fully exploit multimodal information when representations are combined only at the decision level.

When missingness in train affects the imaging modality (\figurename~\ref{fig:ablation_results_fusion}, top right), the differences between fusion strategies become more pronounced. 
The early fusion model starts with performance close to that of the proposed intermediate fusion approach but progressively collapses toward the unimodal baseline as imaging availability is severely reduced, indicating that frozen unimodal representations are insufficient to sustain robust multimodal learning under sparse visual supervision. 
The late fusion model, instead, initially performs similarly to the unimodal baselines and only approaches the performance of the intermediate fusion strategy at higher missing rates (50–75\%), when multimodal supervision is largely diminished and all methods effectively operate under near-unimodal conditions.

\subsection{RQ2: Resilience under Missingness in Test}\label{sec:rq2}
This section examines how the proposed framework performs when deployed under partial or fully unimodal data availability, reflecting clinical conditions where imaging or clinical records may be unavailable at inference time.

\subsubsection{Comparison with Competitors}
Across all testing conditions (\figurename~\ref{fig:results}, bottom panels), our method demonstrates a consistent advantage over the baseline strategies, maintaining resilient performance across the different missing rates for both modalities (see also \tablename~\ref{tab:results}).

When imaging is always available, our approach maintains performance above the unimodal vision encoder for all levels of tabular missingness (\figurename~\ref{fig:results}, bottom left). 
This indicates that multimodal fine-tuning refines the visual representations themselves, enabling the model to benefit from cross-modal interactions even when the auxiliary modality contributes partial information. 
Competitor methods fail to achieve this behavior: the zeros modality composition strategy deteriorates sharply as missingness increases, while model selection is unable to exploit tabular information when it is partially observed, restricting its performance to that of the vision-only model. 
Max pooling performs slightly better than all these baselines and our approach, but remains unable to match the stability and performance of our masked-attention framework on the least informative modality, clinical data.

Indeed, in the complementary scenario where clinical features remain available while imaging is progressively removed (\figurename~\ref{fig:results}, bottom right), performance decreases more markedly. 
This trend reflects the large performance gap between unimodal vision and tabular branches, also visible in their reference lines, which naturally biases multimodal inference toward the more informative imaging modality. 
As a result, when images are fully removed, the model converges to the unimodal tabular performance. 
Nonetheless, our method across all evaluated settings outperforms competitors throughout the degradation regime. 
The zeros modality composition approach once again collapses as missingness increases, while the model-selection baseline is unable to exploit multimodal information when only partial imaging is available, always reaching unimodal performance in the unimodal testing scenario. 
As already noted in the missingness in training scenario, max pooling amplifies the contribution of the most informative modality; however, by neglecting the joint optimization of the unimodal branches, it leads to a suboptimal tabular representation. 
Consequently, performance degrades rapidly as the imaging modality is progressively masked, revealing an over-reliance on visual information and the absence of a balanced fusion strategy.

It is worth noting that such results confirm that the proposed masked-attention mechanism enables robust integration of heterogeneous modalities while avoiding the fragile assumptions embedded in simpler fusion rules.

\subsection{Ablation Analyses}
\paragraph{Competitor-inspired Variants within Our Architecture} 
In the missingness in test scenario (\figurename~\ref{fig:ablation_results_approach}, bottom panels; see also \tablename~\ref{tab:ablation_results}), the observed trends closely mirror those reported in the baseline comparison. 
The zeros modality composition ablation exhibits a steep decline in both stress curves, confirming that constant-value modality composition is incompatible with fine-grained tabular missingness and with the structured nature of visual representations. 
The model-selection variant across all evaluated settings plateaus at unimodal performance whenever one modality becomes unavailable, demonstrating that dynamic switching prevents any meaningful cross-modal interaction under partial observability.

Crucially, neither competitor-inspired variant matches the stability achieved by the proposed masked-attention design. 
These results indicate that robustness does not stem solely from architectural capacity or from strong unimodal encoders, but specifically from explicit attention-level masking combined with modality-dropout, which enables the fusion module to condition its inference exclusively on observed information while remaining exposed to diverse missingness patterns during training.

\paragraph{Alternative Fusion Strategies} 
\figurename~\ref{fig:ablation_results_fusion} (bottom panels) compares the proposed intermediate fusion transformer with its early and late fusion counterparts under missingness in test (see also \tablename~\ref{tab:ablation_results}). 
The limitations of early and late fusion become more evident in this setting. 
The late fusion model exhibits pronounced instability when clinical data is progressively removed (\figurename~\ref{fig:ablation_results_fusion}, bottom left). 
This behavior can be attributed to its reliance on aggregating decision-level outputs from independently trained unimodal predictors, without learning shared representations that account for heterogeneous class distributions. 
As tabular missingness increases, this strategy becomes particularly fragile for underrepresented diagnostic classes, whose predictions rely on weak or noisy signals from a single modality. 
Since both training and evaluation are weighted by class prevalence, small degradations in rare classes can disproportionately affect the weighted AUC, resulting in unstable performance trends. Conversely, when imaging is gradually removed (\figurename~\ref{fig:ablation_results_fusion}, bottom right), performance degrades more smoothly but ultimately converges to the unimodal tabular baseline, highlighting that late fusion neither enhances unimodal representations nor compensates for the loss of the dominant modality.

Across both missingness in test stress curves, the early fusion model across all evaluated settings underperforms the proposed approach. 
Since unimodal encoders remain frozen in this configuration, the model cannot adapt representations to exploit cross-modal interactions, leading to inferior performance and reduced robustness. 
In contrast, the proposed intermediate fusion design, which combines masked attention with joint fine-tuning of unimodal encoders, achieves superior performance and smoother degradation across all missingness regimes.

\subsection{Discussion}

To summarize the main findings and their implications, we highlight six key messages that capture the core contributions of this work.

\begin{enumerate}
\item \textbf{Masked attention with intermediate fusion is the key driver of robustness.}
The results demonstrate that the combination of masked self-attention and intermediate fusion constitutes the central mechanism enabling resilience under modality missingness. By explicitly preventing missing tokens from contributing to both attention and gradient propagation, the model avoids the biases introduced by heuristic strategies and learns representations conditioned exclusively on observed data. This design allows effective cross-modal interaction while maintaining stability across all missingness regimes.

\item \textbf{The framework can be reliably trained on highly unpaired multimodal data.}
The missingness in training experiments (RQ1) show that the model maintains stable performance even when a large fraction of training samples lacks one modality. Unlike competing approaches, the proposed framework leverages both unimodal and partially paired data without discarding incomplete samples or relying on surrogate signals. The synergy between attention-level masking and modality-dropout enables the model to learn consistent representations without overfitting to modality-specific shortcuts, making it particularly suitable for retrospective and multi-center clinical datasets.

\item \textbf{The model adapts dynamically to partial information at inference time.}
Under missingness in test (RQ2), the model exhibits smooth and predictable performance degradation, adapting its predictions to the subset of available modalities. Rather than relying on imputation or model switching, the architecture conditions its reasoning exclusively on observed inputs, effectively mimicking clinical decision-making under partial evidence. This behavior ensures reliable inference across the full spectrum from fully multimodal to strictly unimodal scenarios.

\item \textbf{Heuristic missing-modality strategies are structurally limited.}
Comparative and ablation analyses reveal that commonly used approaches such as constant-value imputation, max pooling, and model selection fail to achieve consistent robustness. These methods either introduce biased representations, ignore cross-modal dependencies, or prevent the model from exploiting partially observed data. In contrast, explicit modeling of missingness at the attention level proves essential to maintain both performance and stability.

\item \textbf{A single model can replace multiple modality-specific systems in practice.}
The consistent performance observed across all missingness conditions indicates that the proposed framework can seamlessly operate under any modality configuration. This eliminates the need for separate unimodal and multimodal models or dynamic selection mechanisms, simplifying deployment and reducing the risk of configuration errors. In clinical settings, this translates into a unified decision-support system capable of handling heterogeneous and incomplete patient data without retraining.

\item \textbf{Learning to condition on observed data is a principled strategy for real-world multimodal AI.}
The results suggest that robustness in multimodal learning does not primarily stem from increased model capacity, but from explicitly constraining the learning process to rely only on available information. This principle enables the model to generalize across heterogeneous and unpredictable missingness patterns, making attention-based multimodal architectures a promising direction for real-world applications where data incompleteness is unavoidable.
\end{enumerate}


\section{Conclusion}\label{sec:conclusion}

In this work, we introduced a multimodal transformer framework designed to perform reliable vision-tabular learning under modality missingness. 
Rather than relying on explicit imputation or heuristic modality switching, the proposed approach conditions its training and inference exclusively on observed information through a combination of masked self-attention and modality-dropout regularization. 
By integrating a vision encoder, a tabular transformer, and a multimodal fusion encoder within a unified intermediate fusion architecture, the model is able to adapt seamlessly to arbitrary subsets of available modalities while preserving stable performance.

We evaluated the proposed framework on the MIMIC-CXR dataset paired with structured clinical data from MIMIC-IV, a setting that naturally reflects real-world healthcare conditions characterized by fragmented modality availability and incomplete labels. 
Through two systematic stress-test protocols, we progressively removed imaging or tabular information either in training or test phases, covering the full spectrum from complete multimodal input to fully unimodal scenarios. 
Across all missingness regimes, the proposed attention-based model outperformed representative baselines from the main families of missing-modality strategies, including modality augmentation, feature space engineering, and model selection approaches.
In particular, our method demonstrated smoother degradation curves and superior resilience whenever one modality was partially or entirely absent, highlighting the advantage of explicit attention-level masking over other approaches. 

Ablation analyses further clarified the source of these gains. 
Employing competitor-inspired configurations within our architectural backbone led to marked performance drops, confirming that resilience does not arise solely from model capacity but from the explicit handling of missingness during representation learning. 
Likewise, comparisons with early and late fusion variants showed that intermediate fusion with joint fine-tuning of unimodal encoders is essential for enabling meaningful cross-modal interactions and for improving unimodal representations themselves. 
Together, these results indicate that the combination of masked attention and modality-dropout constitutes a principled and effective mechanism for learning under heterogeneous and unpredictable data availability.

From a clinical perspective, these properties are particularly relevant.
Crucially, the benefits of the proposed framework emerge already during model training (RQ1), where the missingness in training experiments show that the model can be reliably optimized even when a substantial fraction of the training data is unpaired, with one modality absent for up to $75\%$ of the patients, condition frequently present in retrospective and multi-center clinical dataset. 
Beyond learning from incomplete data, the same architectural principles translate directly into robust behavior at inference time (RQ2), enabling the model to adapt seamlessly to heterogeneous modality availability at deployment. 
Indeed, the proposed framework mirrors the way clinicians reason under partial evidence by dynamically exploiting whichever information is available, without requiring retraining, explicit imputation, or manual model selection. 
The ability to deploy a single model that operates robustly in imaging-only, tabular-only, and multimodal configurations simplifies integration into clinical decision-support pipelines and reduces the risk of brittle behavior caused by incomplete inputs.

Despite these encouraging results, some limitations still remain. 
First, the evaluation was conducted on a single large-scale dataset focused on chest radiography, and generalization to other imaging modalities, clinical tasks, or institutional settings remains to be validated. 
Second, while the proposed attention mechanisms explicitly prevent missing tokens from influencing inference, they do not model the underlying missingness process itself, which could be informative in certain clinical contexts. 
Third, the computational cost associated with transformer-based fusion may pose challenges in resource-constrained deployment scenarios, motivating further investigation into lightweight or sparse-attention variants.

Future work will explore extending the framework to additional modalities, such as clinical text or longitudinal signals, as well as incorporating temporal modeling to handle irregularly sampled patient trajectories. 
Another promising direction is the integration of uncertainty estimation and causal reasoning mechanisms to better capture the clinical implications of missing data. 
Finally, validating the approach across multiple institutions and downstream clinical tasks will be essential to assess its robustness, fairness, and practical impact in real-world healthcare environments.


\section*{Acknowledgment}
This work was partially funded by: 
i) Università Campus Bio-Medico di Roma under the program ``University Strategic Projects'' within the project ``AI-powered Digital Twin for next-generation lung cancEr cAre (IDEA)''; 
ii) from PRIN 2022 MUR 20228MZFAA-AIDA (CUP C53D23003620008); 
iii) from PRIN PNRR 2022 MUR P2022P3CXJ-PICTURE (CUP C53D23009280001);
iv) from PNRR MUR project PE0000013-FAIR;
v) from PNRR-MCNT2-2023-12377755.

Resources are provided by the National Academic Infrastructure for Supercomputing in Sweden (NAISS) and the Swedish National Infrastructure for Computing (SNIC) at Alvis @ C3SE, partially funded by the Swedish Research Council through grant agreements no. 2022-06725 and no. 2018-05973.

\section*{Author Contributions}
\textbf{Camillo Maria Caruso:} Conceptualization, Data curation, Formal analysis, Investigation, Methodology, Software, Validation, Visualization, Writing – original draft, Writing – review \& editing;
\textbf{Valerio Guarrasi:} Conceptualization, Formal analysis, Funding acquisition, Investigation, Methodology, Project administration, Resources, Supervision, Validation, Visualization, Writing – review \& editing;
\textbf{Paolo Soda:} Conceptualization, Formal analysis, Funding acquisition, Investigation, Methodology, Project administration, Resources, Supervision, Validation, Visualization, Writing – review \& editing.



\bibliographystyle{IEEEtran} 
\bibliography{bibliography.bib}

\appendix
\clearpage
\onecolumn

\section{Dataset Preparation Details}\label{app:data}

We use a multimodal dataset obtained by pairing chest X-Rays from the MIMIC-CXR repository with structured clinical information from MIMIC-IV. 
The initial imaging collection consisted of 377110 chest radiographs corresponding to 227835 studies. 
To ensure consistency between modalities, we remove any study associated with a subject identifier not present in MIMIC-IV, as well as duplicate studies referring to the same patient. 
This procedure also prevented potential data leakage across training and testing phases. After this alignment step, each radiograph could be associated with the patient's age and sex.

Temporal matching was then performed to link each imaging study to the corresponding emergency department stay. 
Using the study timestamp, we identified the admission interval for the same subject. 
Studies for which no valid association could be established were excluded. 
This matching enabled the extraction of vital signs recorded during triage, including temperature, heart rate, respiration rate, oxygen saturation, systolic and diastolic blood pressures, as well as a short text description of the patient's presenting problem (\textit{description} feature in \tablename~\ref{tab:vital_signs}).

Because triage measurements may be temporally distant from the imaging acquisition, we introduce an additional refinement step that, for each study, searches within the stay information for vital-sign measurements closer in time to the radiograph than those obtained at triage. 
When such measurements were available, the corresponding values replaced the initial triage-derived ones, ensuring a more reliable temporal alignment between clinical and imaging information.

A detailed inspection of the extracted measurements revealed the presence of physiologically implausible values (e.g., temperatures of several thousand degrees). 
To mitigate the effect of such errors, we applied two different outlier removal strategies. 
First, for temperature, heart rate, respiration rate, and oxygen saturation, we exclude values falling outside admissible physiological ranges derived from the literature~\cite{bib:vital_sign}. 
Second, for systolic and diastolic blood pressures, where such ranges were not readily available, we use the interquartile range (IQR) method, removing values falling outside the interval $[Q_1 - 1.5\cdot\mathrm{IQR}$; $Q_3 + 1.5\cdot\mathrm{IQR}]$, where $IQR = Q_3 - Q_1$, $Q_1$ and $Q_3$ are the first and third quartiles, respectively.
It is worth noting that removed values are considered as missing data in our approach.
Table~\ref{tab:vital_signs} further describes the clinical features included in the study, reporting their missing rates and the admissible ranges used during preprocessing. 

The resulting multimodal dataset contains 62071 imaging-clinical pairs, consisting of 1 scan and 10 clinical features for each patient. 

\begin{table}[ht]
 \centering
 \scalebox{.9}{
 \begin{tabular}{c|c|c}
 \toprule
 \textbf{Feature} & \textbf{Missingness} & \textbf{Range}\\
 \midrule
 Age & 0.00\% & - \\
 Sex & 0.00\% & - \\
 Ethnicity & 1.25\% & - \\
 Temperature & 35.59\% & [86-113] $^\circ$F \\
 Heart Rate & 3.17\% & [25-225] bpm \\
 Respiration Rate & 4.06\% & [7-40] brpm \\
 Oxygen Saturation & 8.20\% & [50-120] \% \\
 Systolic Blood Pressure & 5.17\% & [69.5-185.5] mmHg \\
 Diastolic Blood Pressure & 4.97\% & [34.5-110.5] mmHg \\
 Description & 50.41\% & - \\
 \bottomrule
 \end{tabular}}
 \caption{List of clinical features used, their missing rates and the ranges used to avoid extreme outliers that are physiologically irrelevant.}
 \label{tab:vital_signs}
\end{table}

\section{Complete Results}\label{app:results}

This appendix reports the complete numerical results underlying the stress-test analyses presented in sections~\ref{sec:rq1} and \ref{sec:rq2}. 
\tablename s~\ref{tab:results_train}-\ref{tab:ablation_results} provide a comprehensive breakdown of the weighted AUC scores obtained under controlled modality-missingness conditions, spanning both missingness in training and in test scenarios, and enabling precise inspection of model behavior beyond the aggregated performance curves.

\tablename s~\ref{tab:results_train} and \ref{tab:results} compare the proposed masked-attention multimodal framework with representative competitor strategies under increasing modality missingness, following the two experimental protocols described in section~\ref{sec:experiments}.
Specifically, \tablename~\ref{tab:results_train} reports results obtained when modality missingness is introduced during training, while the test set remains fully multimodal. 
For each modality (imaging and tabular), the missing rate is progressively increased from $0\%$ to $75\%$, and performance is evaluated on complete multimodal inputs. 
This setting assesses the ability of each method to learn robust multimodal representations from highly unpaired training data, where a substantial fraction of patients lacks one modality.

Conversely, \tablename~\ref{tab:results} reports results obtained under modality missingness in test, where all models are trained on fully available multimodal data and missingness is introduced only at inference time. 
In this scenario, the missing rate spans from $0\%$ to $100\%$, with the $100\%$ condition corresponding to fully unimodal inference. 
This protocol evaluates deployment-time robustness and quantifies how gracefully each approach degrades as information is progressively removed.

\tablename s~\ref{tab:ablation_results_train} and \ref{tab:ablation_results} focus on ablation analyses within the proposed architecture, isolating the contribution of different design choices under the same missingness regimes.
In particular, \tablename~\ref{tab:ablation_results_train} reports the missingness in training scenario, where one modality is progressively masked during training ($0\%$-$75\%$), while testing is always performed on fully multimodal data. 
The proposed model is compared against ablation variants that incorporate competitor-inspired strategies (zeros modality composition and model-selection) or alternative fusion mechanisms (early and late fusions), allowing the assessment of how explicit attention-level masking and intermediate fusion contribute to stable learning from incomplete data.
Finally, \tablename~\ref{tab:ablation_results} reports the missingness in test ablation results, where all models are trained on fully paired multimodal data and missingness is applied only at inference time ($0\%$-$100\%$). 
This table highlights how different architectural choices affect resilience to missing data at deployment, and clarifies the role of masked self-attention and joint multimodal fine-tuning in enabling the model to condition its predictions exclusively on observed modalities without collapsing toward brittle unimodal baselines.

For each missingness condition, the best-performing method is highlighted in bold, while the second-best performance is highlighted in blue, with ties treated across all evaluated settings. 
Together, these four tables provide a complete quantitative characterization of model performance across all missingness configurations considered in this work, complementing the stress-test plots and supporting the conclusions drawn in section~\ref{sec:results} regarding resilience, stability, and effective multimodal integration under pervasive modality incompleteness.
Indeed, we can easily note that, across the majority of missingness configurations and experimental settings, the proposed method across all evaluated settings ranks among the top two performing approaches, highlighting its ability to maintain competitive performance and resilience even under severe and heterogeneous modality-missingness conditions.

\begin{table}[ht]
 \centering
 \resizebox{\textwidth}{!}{
 \begin{tabular}{c|c|c|c|c||c|c|c|c}
 \toprule
 & \multicolumn{4}{c||}{\textbf{Tabular Missingness Results}} & \multicolumn{4}{c}{\textbf{Imaging Missingness Results}} \\
 & \multicolumn{4}{c||}{\textbf{Train Missing Rate}} & \multicolumn{4}{c}{\textbf{Train Missing Rate}} \\
 \textbf{Approach} & 0\% & 25\% & 50\% & 75\% & 0\% & 25\% & 50\% & 75\% \\
 \midrule
 Ours & \textbf{82.58$\pm$0.26} & \textbf{82.88$\pm$0.31} & \textbf{82.60$\pm$0.24} & \bl\textbf{82.24$\pm$0.28}\bb & \textbf{82.58$\pm$0.26} & \textbf{81.60$\pm$0.40} & \bl\textbf{81.75$\pm$0.35}\bb & \textbf{80.57$\pm$0.74} \\
 Zeros & 82.00$\pm$0.47 & 81.37$\pm$0.93 & 72.46$\pm$5.74 & 67.34$\pm$7.11 & 82.00$\pm$0.47 & 78.93$\pm$1.18 & 78.84$\pm$1.22 & 75.09$\pm$1.60 \\
 Max Pooling & \bl\textbf{82.25$\pm$0.32}\bb & \bl\textbf{82.38$\pm$0.37}\bb & \bl\textbf{81.89$\pm$0.43}\bb & \textbf{82.48$\pm$0.37} & \bl\textbf{82.25$\pm$0.32}\bb & 80.44$\pm$0.58 & 78.05$\pm$1.16 & 57.76$\pm$4.76 \\
 Model Selection & 82.00$\pm$0.47 & 81.42$\pm$0.63 & 75.91$\pm$6.48 & 81.19$\pm$0.78 & 82.00$\pm$0.47 & \bl\textbf{80.75$\pm$0.69}\bb & \textbf{82.37$\pm$0.36} & \bl\textbf{78.47$\pm$0.78}\bb \\
 \midrule
 Unimodal & 72.27$\pm$0.47 & 72.01$\pm$1.05 & 68.97$\pm$1.54 & 71.50$\pm$0.61 & 78.38$\pm$0.51 & 79.37$\pm$0.62 & 79.75$\pm$0.81 & 77.10$\pm$0.79 \\
 \bottomrule
 \end{tabular}}
 \caption{Mean weighted AUC (mean $\pm$ standard error over five cross-validation folds) obtained under progressively increasing train-time modality missingness. Results are reported separately for tabular-missingness (left block), where clinical features are masked while imaging remains available, and imaging-missingness (right block), where chest radiographs are masked while tabular data remain available. The proposed model is compared against representative missing-modality strategies, including zero-imputation, max pooling, and model selection. Best results for each missingness level are shown in bold, with second-best results highlighted in blue.}
 \label{tab:results_train}
\end{table}

\begin{table}[ht]
 \centering
 \resizebox{\textwidth}{!}{
 \begin{tabular}{c|c|c|c|c|c||c|c|c|c|c}
 \toprule
 & \multicolumn{5}{c||}{\textbf{Tabular Missingness Results}} & \multicolumn{5}{c}{\textbf{Imaging Missingness Results}} \\
 & \multicolumn{5}{c||}{\textbf{Test Missing Rate}} & \multicolumn{5}{c}{\textbf{Test Missing Rate}} \\
 \textbf{Approach} & 0\% & 25\% & 50\% & 75\% & 100\% & 0\% & 25\% & 50\% & 75\% & 100\% \\
 \midrule
 Ours & \textbf{82.58$\pm$0.26} & \bl\textbf{81.13$\pm$0.32}\bb & \bl\textbf{80.29$\pm$0.39}\bb & \bl\textbf{79.94$\pm$0.36}\bb & \bl\textbf{80.11$\pm$0.21}\bb & \textbf{82.58$\pm$0.26} & \textbf{77.12$\pm$0.68} & \textbf{73.80$\pm$0.90} & \textbf{71.92$\pm$0.68} & \bl\textbf{71.83$\pm$0.69}\bb \\
 Zeros & 82.00$\pm$0.47 & 77.37$\pm$1.42 & 73.82$\pm$2.86 & 70.40$\pm$4.48 & 68.25$\pm$6.00 & 82.00$\pm$0.47 & \bl\textbf{76.36$\pm$1.36}\bb & \bl\textbf{72.63$\pm$1.77}\bb & 69.83$\pm$1.61 & 68.95$\pm$0.91\\
 Max Pooling & \bl\textbf{82.25$\pm$0.32}\bb & \textbf{81.76$\pm$0.30} & \textbf{81.34$\pm$0.30} & \textbf{80.99$\pm$0.29} & \textbf{80.75$\pm$0.31} & \bl\textbf{82.25$\pm$0.32}\bb & 76.10$\pm$1.54 & 71.40$\pm$2.26 & 67.40$\pm$2.43 & 64.55$\pm$2.25 \\
 Model Selection & 82.00$\pm$0.47 & 80.41$\pm$0.33 & 79.27$\pm$0.36 & 78.56$\pm$0.38 & 78.38$\pm$0.51 & 82.00$\pm$0.47 & 75.99$\pm$0.72 & 72.56$\pm$1.02 & \bl\textbf{71.31$\pm$0.77}\bb & \textbf{72.27$\pm$0.47}\\
 \midrule
 Unimodal & - & - & - & - & 78.38$\pm$0.51 & - & - & - & - & \textbf{72.27$\pm$0.47} \\
 \bottomrule
 \end{tabular}}
 \caption{Mean weighted AUC (mean $\pm$ standard error over five cross-validation folds) obtained under progressively increasing modality missingness in test. Results are reported separately for tabular-missingness (left block), where clinical features are masked while imaging remains available, and imaging-missingness (right block), where chest radiographs are masked while tabular data remain available. The proposed model is compared against representative missing-modality strategies, including zero-imputation, max pooling, and model selection. The unimodal reference performance is reported at 100\% missingness for each modality. Best results for each missingness level are shown in bold, with second-best results highlighted in blue.}
 \label{tab:results}
\end{table}

\begin{table}[ht]
 \centering
 \resizebox{\textwidth}{!}{
 \begin{tabular}{c|c|c|c|c||c|c|c|c}
 \toprule
 & \multicolumn{4}{c||}{\textbf{Tabular Missingness Results}} & \multicolumn{4}{c}{\textbf{Imaging Missingness Results}} \\
 & \multicolumn{4}{c||}{\textbf{Train Missing Rate}} & \multicolumn{4}{c}{\textbf{Train Missing Rate}} \\
 \textbf{Approach} & 0\% & 25\% & 50\% & 75\% & 0\% & 25\% & 50\% & 75\% \\
 \midrule
 Ours & \textbf{82.58$\pm$0.26} & 82.88$\pm$0.31 & \textbf{82.60$\pm$0.24} & \bl\textbf{82.24$\pm$0.28}\bb & \textbf{82.58$\pm$0.26} & \textbf{81.60$\pm$0.40} & \bl\textbf{81.75$\pm$0.35}\bb & \bl\textbf{80.57$\pm$0.74}\bb \\
 Zeros & 82.30$\pm$0.31 & \textbf{82.94$\pm$0.14} & 82.42$\pm$0.29 & 82.07$\pm$0.40 & 82.30$\pm$0.31 & 81.09$\pm$0.39 & 81.17$\pm$0.55 & 79.36$\pm$1.82 \\
 Model Selection & \bl\textbf{82.40$\pm$0.27}\bb & \bl\textbf{82.90$\pm$0.16}\bb & \bl\textbf{82.57$\pm$0.35}\bb & \bl\textbf{82.24$\pm$0.36}\bb & \bl\textbf{82.40$\pm$0.27}\bb & \bl\textbf{81.41$\pm$0.89}\bb & \textbf{82.42$\pm$0.42} & 78.44$\pm$0.94 \\
 \midrule
 Early Fusion & 81.91$\pm$0.47 & 82.69$\pm$0.32 & 81.75$\pm$0.72 & \textbf{82.45$\pm$0.22} & 81.91$\pm$0.47 & 81.09$\pm$0.61 & 79.83$\pm$0.75 & 77.11$\pm$0.71 \\
 Late Fusion & 78.98$\pm$0.32 & 79.13$\pm$0.30 & 78.43$\pm$0.79 & 78.35$\pm$0.42 & 78.98$\pm$0.32 & 80.84$\pm$0.98 & 81.71$\pm$0.48 & \textbf{80.87$\pm$0.52} \\
 \midrule
 Unimodal & 72.27$\pm$0.47 & 72.01$\pm$1.05 & 68.97$\pm$1.54 & 71.50$\pm$0.61 & 78.38$\pm$0.51 & 79.37$\pm$0.62 & 79.75$\pm$0.81 & 77.10$\pm$0.79 \\
 \bottomrule
 \end{tabular}}
 \caption{Mean weighted AUC (mean ± standard error over five cross-validation folds) for the ablation analysis under progressively increasing train-time modality missingness. Results are reported separately for tabular-missingness (left block), where clinical features are masked while imaging remains available, and imaging-missingness (right block), where chest radiographs are masked while tabular data remain available. The proposed model is compared with competitor-inspired variants (zero-imputation and model selection) and alternative fusion strategies (early fusion and late fusion). Best results for each missingness level are shown in bold, with second-best results highlighted in blue.}
 \label{tab:ablation_results_train}
\end{table}

\begin{table}[ht]
 \centering
 \resizebox{\textwidth}{!}{
 \begin{tabular}{c|c|c|c|c|c||c|c|c|c|c}
 \toprule
 & \multicolumn{5}{c||}{\textbf{Tabular Missingness Results}} & \multicolumn{5}{c}{\textbf{Imaging Missingness Results}} \\
 & \multicolumn{5}{c||}{\textbf{Test Missing Rate}} & \multicolumn{5}{c}{\textbf{Test Missing Rate}} \\
 \textbf{Approach} & 0\% & 25\% & 50\% & 75\% & 100\% & 0\% & 25\% & 50\% & 75\% & 100\% \\
 \midrule
 Ours & \textbf{82.58$\pm$0.26} & \textbf{81.13$\pm$0.32} & \textbf{80.29$\pm$0.39} & \textbf{79.94$\pm$0.36} & \textbf{80.11$\pm$0.21} & \textbf{82.58$\pm$0.26} & \bl\textbf{77.12$\pm$0.68}\bb & 73.80$\pm$0.90 & 71.92$\pm$0.68 & \bl\textbf{71.83$\pm$0.69}\bb \\
 Zeros & \bl\textbf{82.30$\pm$0.31}\bb & 74.91$\pm$1.80 & 68.13$\pm$3.27 & 61.65$\pm$4.87 & 55.80$\pm$6.64 & \bl\textbf{82.30$\pm$0.31}\bb & 76.24$\pm$1.41 & 72.30$\pm$1.74 & 69.87$\pm$1.21 & 69.42$\pm$0.75 \\
 Model Selection & \textbf{82.58$\pm$0.26} & 80.41$\pm$0.63 & 78.98$\pm$0.73 & 78.37$\pm$0.49 & 78.38$\pm$0.51 & \textbf{82.58$\pm$0.26} & \textbf{77.23$\pm$0.72} & \bl\textbf{73.95$\pm$0.95}\bb & \bl\textbf{72.22$\pm$0.75}\bb & \textbf{72.27$\pm$0.47} \\
 \midrule
 Early Fusion & 81.91$\pm$0.47 & \bl\textbf{80.41$\pm$0.48}\bb & \bl\textbf{79.38$\pm$0.52}\bb & \bl\textbf{78.84$\pm$0.54}\bb & \bl\textbf{78.88$\pm$0.59}\bb & 81.91$\pm$0.47 & 75.72$\pm$0.55 & 72.18$\pm$0.83 & 70.79$\pm$0.61 & \bl\textbf{71.83$\pm$0.69}\bb \\
 Late Fusion & 78.98$\pm$0.32 & 74.16$\pm$0.75 & 72.15$\pm$0.98 & 73.57$\pm$0.84 & 78.38$\pm$0.51 & 78.98$\pm$0.32 & 76.16$\pm$0.44 & \textbf{74.29$\pm$0.42} & \textbf{72.90$\pm$0.38} & \textbf{72.27$\pm$0.47}\\
 \midrule
 Unimodal & - & - & - & - & 78.38$\pm$0.51 & - & - & - & - & \textbf{72.27$\pm$0.47} \\
 \bottomrule
 \end{tabular}}
 \caption{Mean weighted AUC (mean ± standard error over five cross-validation folds) for the ablation analysis under progressively increasing modality missingness in test. Results are reported separately for tabular-missingness (left block), where clinical features are masked while imaging remains available, and imaging-missingness (right block), where chest radiographs are masked while tabular data remain available. The proposed model is compared with competitor-inspired variants (zero-imputation and model selection) and alternative fusion strategies (early fusion and late fusion). The unimodal reference performance is reported at 100\% missingness for each modality. Best results for each missingness level are shown in bold, with second-best results highlighted in blue.}
 \label{tab:ablation_results}
\end{table}

\end{document}